\documentclass[11pt]{article}
\usepackage{listings}
\lstdefinestyle{pseudocode}{language=Python, basicstyle=\ttfamily\small, keywordstyle=\color{blue}, commentstyle=\color{gray}, stringstyle=\color{red}, showstringspaces=false, breaklines=true}
\usepackage{microtype}
\usepackage{graphicx}
\usepackage{booktabs}
\usepackage{amsmath}
\usepackage{amsfonts}

\usepackage{amssymb}
\usepackage{bm}
\usepackage{enumitem}
\usepackage{xspace}
\usepackage{makecell}
\usepackage{multirow}
\usepackage{algorithm}
\usepackage{algpseudocode}
\usepackage{arxiv}
\usepackage{subcaption}

\usepackage{hyperref}

\newcommand{\methodname}{HLA}
\usepackage{url}
\usepackage{caption}

\newcommand{\reporttitle}{HLA: Expressive Hybrid Linear Attention via Chunk-Wise Dynamic Mixing}
\newcommand{\reportshorttitle}{Short Title}
\newcommand{\reportauthors}{
\textbf{Zhuokun Chen}\textsuperscript{\normalfont 1}
\quad
\textbf{Xi Lin}\textsuperscript{\normalfont 2} \quad
\textbf{Xiyu Wu}\textsuperscript{\normalfont 2} \quad
\textbf{Jiahao He}\textsuperscript{\normalfont 2}  \\
\textbf{Jianfei Cai}\textsuperscript{1} \quad
\textbf{Bohan Zhuang}\textsuperscript{2}
}
\newcommand{\reportaffiliations}{\textsuperscript{1}Monash University \quad
\textsuperscript{2}Zhejiang University}

\newcommand{\reportproject}{\href{https://caesarhhh.github.io/hla/}{https://caesarhhh.github.io/hla/}}
\newcommand{\reportemail}{\href{mailto:bohan.zhuang@zju.edu.cn}{bohan.zhuang@zju.edu.cn}}
\newcommand{\reportorcid}{\href{https://orcid.org/0000-0000-0000-0000}{First Author}}
\newcommand{\reportkeywords}{Linear Attention, Long-Context Modeling, Efficient Inference, Gated DeltaNet}

\newcommand{\reportpdfauthors}{First Author, Second Author, Third Author}
\newcommand{\reportlogoheight}{0.62cm}

\techreportlabel{ZIP Lab Technical Report}
\techreportshorttitle{\reportshorttitle}
\techreportsetlogoheight{\reportlogoheight}
\techreportproject{\reportproject}
\techreportemail{\reportemail}
\techreportorcid{\reportorcid}
\techreportkeywords{\reportkeywords}

\techreportlogos{%
  \techreportlogo{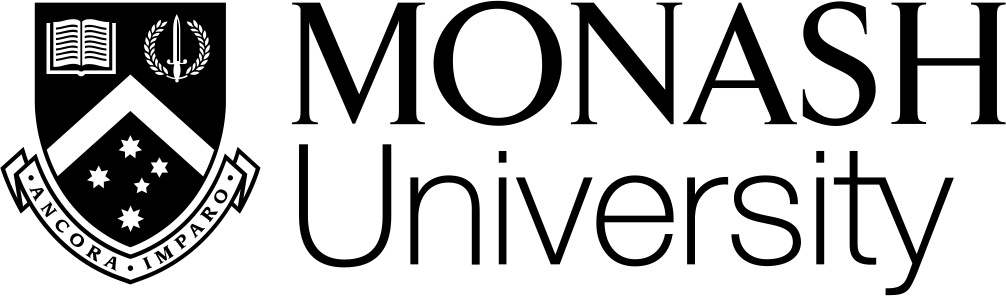}
  \techreportlogo{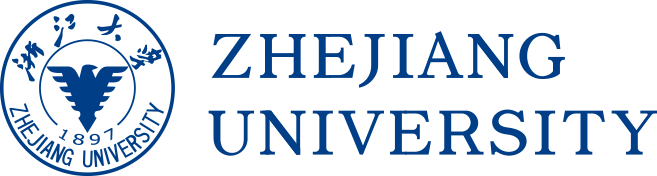}
}

\hypersetup{
  colorlinks=true,
  linkcolor=ziplabblue,
  citecolor=ziplabblue,
  urlcolor=ziplabblue,
  pdftitle={\reporttitle},
  pdfauthor={\reportpdfauthors},
  pdfsubject={ZIP Lab technical report},
  pdfkeywords={\reportkeywords}
}

\begin{document}

\ziplabmaketitle{\reporttitle}{\reportauthors}{\reportaffiliations}{%
  \begin{center}
  \end{center}
}
\ziplabendtitle{Linear attention enables efficient long-context autoregressive decoding by
compressing history into recurrent states, but this compression can make
selective access to sparse and distant information difficult. Existing
chunk-based extensions increase memory capacity, yet learned chunk-mixing
coefficients may remain fixed with respect to input content and therefore
cannot adapt historical access to each query. We introduce
\emph{Hybrid Linear Attention} (HLA), a query-dependent
chunk-level attention mechanism for Gated DeltaNet (GDN).
HLA represents each completed chunk as an exact affine state
transition and computes content-dependent routing gates from compact,
self-attentively pooled representatives. Each gate interpolates the
corresponding historical transition with the identity map, controlling
both the chunk's additive memory and its transformation of earlier
states. Effective-support regularization further encourages concentrated
routing for sparse inference.
We evaluate HLA under both pretrained adaptation and
from-scratch training. Across Qwen3.5 models from 0.8B to 9B,
HLA consistently improves over native GDN and fixed chunk
mixing, with gains of up to 5.57 percentage points on LongBench-V2 and
3.97 points on RULER. In a controlled from-scratch 1.3B setting trained
for 100B tokens with a 4K context, HLA also improves RULER
performance from 4K to 32K, with gains increasing from 0.83 points at
4K to 4.22 points at 32K. These results demonstrate that query-dependent
composition of recurrent memory improves long-context modeling and
remains effective beyond the training context while using compact
per-chunk affine summaries.
}

\section{Introduction}
Softmax attention is the standard mechanism for long-context sequence modeling and remains highly effective for content-dependent retrieval~\citep{vaswani2017attention,brown2020language,chowdhery2023palm}. However, in autoregressive generation its KV cache grows with context length, which makes very long-context inference increasingly memory-intensive~\citep{kwon2023efficient,dao2022flashattention,zaheer2020big,dai2019transformer,zhang2023h2o,liu2023scissorhands,li2024snapkv,liu2023scissorhands,ainslie2023gqa,liu2024minicache,tay2020sparse}.

Linear attention offers a different trade-off~\citep{dao2024transformers,yang2024gated,zhang2026mhla,munkhdalai2024leave}. By removing the nonlinearity operation of softmax, it allows the model to first aggregate key-value interactions into a fixed-size matrix whose dimension does not grow with context length, thereby avoiding the memory-bound behavior of softmax attention in long contexts~\citep{schlag2021linear}. However, this efficiency comes at the cost of limited expressivity~\citep{zaheer2020big,beltagy2020longformer}. In particular, the rank of linear attention is bounded by the feature dimension, which restricts its ability to represent sharp and highly selective attention patterns when the context becomes very long~\citep{xiong2021nystromformer}. One useful way to understand this gap is that softmax attention can produce much more concentrated attention distributions, allowing the model to focus strongly on a small subset of relevant context, whereas linear-attention variants often distribute their scores more evenly over the visible history~\citep{yang2024gated,zhang2026mhla}. This difference becomes increasingly important when relevant evidence is sparse and must be selected from a large amount of irrelevant context~\citep{lewis2020retrieval}.

\begin{figure*}[t]
  \centering

  \begin{subfigure}[t]{0.34\linewidth}
    \centering
    \includegraphics[width=\linewidth]{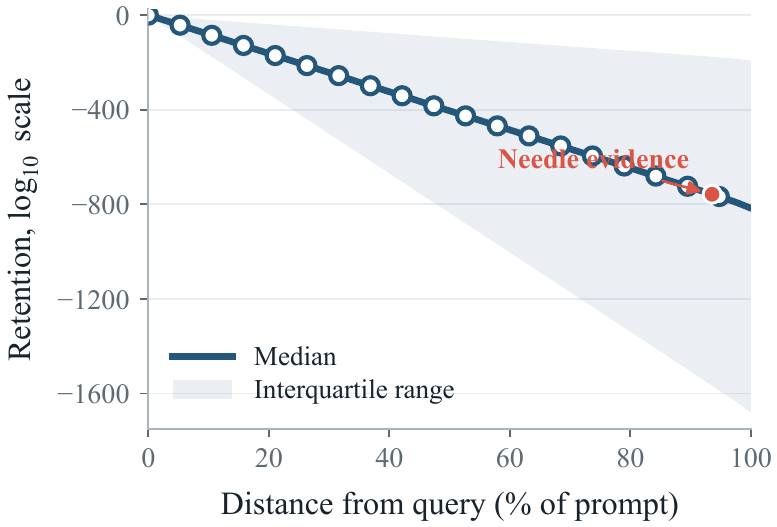}
    \caption{Cumulative recurrence retention in GLA.}
    \label{fig:intro-gla-retention}
  \end{subfigure}
  \hfill
  \begin{subfigure}[t]{0.335\linewidth}
    \centering
    \includegraphics[width=\linewidth]{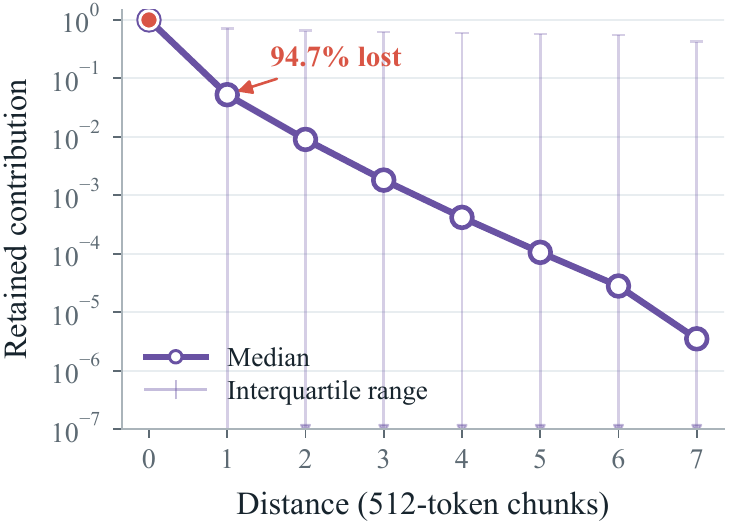}
    \caption{Long-range contribution retention in GDN.}
    \label{fig:intro-gdn-retention}
  \end{subfigure}
  \hfill
  \begin{subfigure}[t]{0.31\linewidth}
    \centering
    \includegraphics[width=\linewidth]{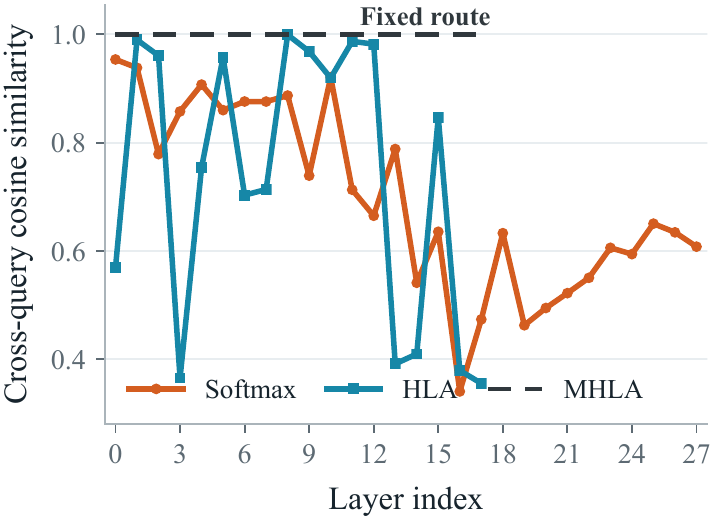}
    \caption{Cross-sample chunk-level similarity.}
    \label{fig:intro-query-dependence}
  \end{subfigure}

  \caption{
  \textbf{Motivating observations for selective long-context retrieval.}
  (a) In a pretrained GLA model, the cumulative recurrent retention of a
state written at each historical position decreases with its distance to
the final query position in a 4K-token prompt. The curve reports the
median across layers and gate dimensions, with the shaded region denoting
the interquartile range (IQR).
  (b) In pretrained GDN, the affine-state contribution of the chunk containing an early needle is progressively attenuated as it propagates through subsequent 512-token chunks. The curve reports the median normalized contribution over layer--head pairs, with IQR shading.
  (c) Chunk-level relevance can vary substantially across inputs. For a selected low-similarity LongBench pair, we report, at each layer, the minimum cross-sample cosine similarity over heads for softmax attention and \methodname{}; MHLA remains at one because its chunk weights are input-independent. Together, these observations motivate query-dependent chunk-level attention for preserving and selectively accessing long-range information.
  }
  \label{fig:intro-insights}
\end{figure*}

Recent work has made substantial progress toward alleviating this limitation through improved memory compression, retrieval, and recurrent state design~\citep{raecompressive,wumemorizing,yang2024gated,yanggated,zhang2026mhla,choromanskirethinking}. Gated Linear Attention (GLA)~\citep{yang2024gated} uses data-dependent gating to regulate recurrent memory, while GDN~\citep{yanggated} further introduces delta-rule updates for more selective state modification. MHLA~\citep{zhang2026mhla} instead partitions the sequence into chunks, maintains a separate linear state for each, and combines them with fixed learned weights to increase the effective capacity of linear attention. However, two limitations remain. (1) Sparse information stored early in the sequence can still be progressively attenuated as the context grows. In GLA, each historical state write is repeatedly modulated by subsequent recurrent gates; as shown in Figure~\ref{fig:intro-insights}(a), the cumulative recurrence retention of early source positions, including the inserted needle, is substantially lower than that of more recent positions in a 4K-token prompt. GDN provides more selective memory updates, but early information must still propagate through a long chain of subsequent affine state transitions and can therefore be progressively attenuated or overwritten. Figure~\ref{fig:intro-insights}(b) illustrates this effect: the contribution of the needle-containing chunk decreases by several orders of magnitude as additional chunks are processed. (2) The relevance of historical context is inherently input- and query-dependent~\citep{liu2024lost}, whereas fixed chunk mixing cannot adapt to such variation. Figure~\ref{fig:intro-insights}(c) illustrates this issue using a low-similarity pair selected from LongBench~\citep{bai2024longbench}: both softmax attention and \methodname{} exhibit substantially different chunk-level patterns across the two inputs, while MHLA applies the same input-independent weights and therefore yields a cross-sample cosine similarity of one. These observations suggest that effective long-context retrieval requires query-dependent mechanisms that can preserve sparse historical information and selectively control its influence on the current query.


In this work, we introduce \emph{Hybrid Linear Attention} (\methodname),
which augments Gated DeltaNet with lightweight, query-dependent
chunk-level routing. The key idea is to make the influence of historical
chunks depend explicitly on the current query. We exploit the affine
structure of GDN to summarize each completed chunk as an exact state
transition with a multiplicative transformation and an additive memory
term. For each query token, \methodname{} predicts chunk-level routing
gates from a small set of self-attentively pooled representatives. Rather
than using these gates only to combine chunk readouts, \methodname{}
applies them to the complete GDN transitions by interpolating each
historical transition with the identity map. This allows the query to
control both the memory introduced by a chunk and how that chunk
transforms earlier information, making the composition of historical GDN
transitions explicitly query-dependent.

We evaluate \methodname{} on retrieval-intensive long-context language modeling tasks, with a particular focus on accessing sparse and distant information over long sequences. \methodname{} requires only lightweight adaptation of a pretrained GDN-based model, while directly enhancing the long-range retrieval capability of its recurrent layers.

Our contributions are summarized as follows:
\begin{itemize}
\item We identify a key limitation of recurrent linear attention for long-context retrieval: although modern architectures such as GDN employ content-dependent memory updates, sparse information written early in the sequence can still be progressively attenuated or overwritten as it passes through a long chain of subsequent state transitions.

\item We propose \methodname{}, which augments GDN with query-dependent chunk-level attention over exact chunk-wise affine transitions. The resulting attention weights selectively control the complete transition of each historical chunk, including both its additive memory and its transformation of earlier states.

\item We develop a practical implementation of query-dependent transition
composition and evaluate HLA under both pretrained adaptation and
from-scratch training. Across LongBench-V2 and RULER, HLA improves
long-context performance over native GDN and fixed chunk mixing, including
generalization beyond the training context.

\end{itemize}

\section{Related Work}

\textbf{Linear attention and state-based models.}~
An alternative line of work replaces dense token-level global attention with recurrent or linear-time state updates~\citep{peng2023rwkv,dai2019transformer}. Early linear-attention formulations showed that autoregressive Transformers can be rewritten as recurrent models with additive state updates, yielding linear-time decoding and bounded memory growth~\citep{katharopoulos2020transformers}. Linear attention provides the basic formulation: instead of storing all past keys and values explicitly, it accumulates additive key-value statistics in a compressed state whose size does not grow with sequence length. This perspective is also closely connected to the structured-state-space view of sequence modeling developed in the SSM-duality literature and to recent selective state-space models such as Mamba~\citep{dao2024transformers,gu2023mamba} and HiPPO~\citep{gu2020hippo}. Recent models build on this idea in different ways, but each also exposes limitations that motivate our design. Gated Linear Attention (GLA) improves state-based modeling through gated recurrent updates, which help suppress redundant history and stabilize memory dynamics~\citep{yang2024gated}. However, GLA still relies on a shared compressed state, so its long-range weighting patterns often remain relatively diffuse and less selective than those of softmax attention when relevant evidence is sparse. GDN further strengthens recurrent memory through data-dependent decay and delta-rule updates~\citep{yanggated}, allowing the model to selectively forget and overwrite historical information. Nevertheless, once a past segment has been incorporated into the recurrent trajectory, its effect on future states is determined before future queries are observed. Information written early in the sequence must also pass through a long chain of subsequent state transitions, making it vulnerable to progressive attenuation or overwriting as the context grows. MHLA~\citep{zhang2026mhla} takes a complementary approach by maintaining multiple chunk-level linear states to increase representational capacity, but aggregates them using a shared, fixed weighting scheme, which limits its ability to adapt retrieval to individual queries. In all of these methods, the central difficulty is similar: efficiently compressed history remains difficult to access in a sharp, query-dependent manner when relevant evidence is sparse and distant. Our method is most closely related to this family of efficient long-context models, but differs by decomposing history into exact chunk-level state transitions and making their composition explicitly query-dependent.

\textbf{Hybrid attention models.}~Recent hybrid linear-attention architectures increasingly follow a common pattern: most layers use linear or recurrent attention for long-context efficiency, while a small number of layers retain full softmax attention for precise retrieval and information mixing. Recent hybrid architectures combine recurrent linear-attention layers
with a smaller number of softmax-attention layers to balance efficiency
and retrieval quality~\citep{yang2024gated,ai2025kimi,team2025ring}. Kimi Linear~\citep{ai2025kimi} adopts a more explicit layerwise recipe, interleaving KDA-based linear attention with full-attention MLA blocks in a 3:1 ratio to retain strong retrieval behavior while reducing overall decoding cost. The Ring-linear series~\citep{team2025ring} from Ling Team also studies this design space directly, explicitly mixing linear attention and softmax attention and analyzing how their layerwise ratio affects the performance of hybrid models. These architectures achieve a trade-off between memory efficiency and modeling quality by combining predominantly linear or recurrent layers with a small number of softmax-attention layers. However, the linear layers themselves still compress long histories into recurrent states and therefore remain susceptible to losing or attenuating sparse long-range information as the context grows. In contrast, our method augments GDN with an additional lightweight, query-dependent chunk-level attention mechanism, directly improving selective access to historical information within the linear-attention layers themselves.

\begin{figure}[t]
    \centering
    \includegraphics[width=\linewidth]{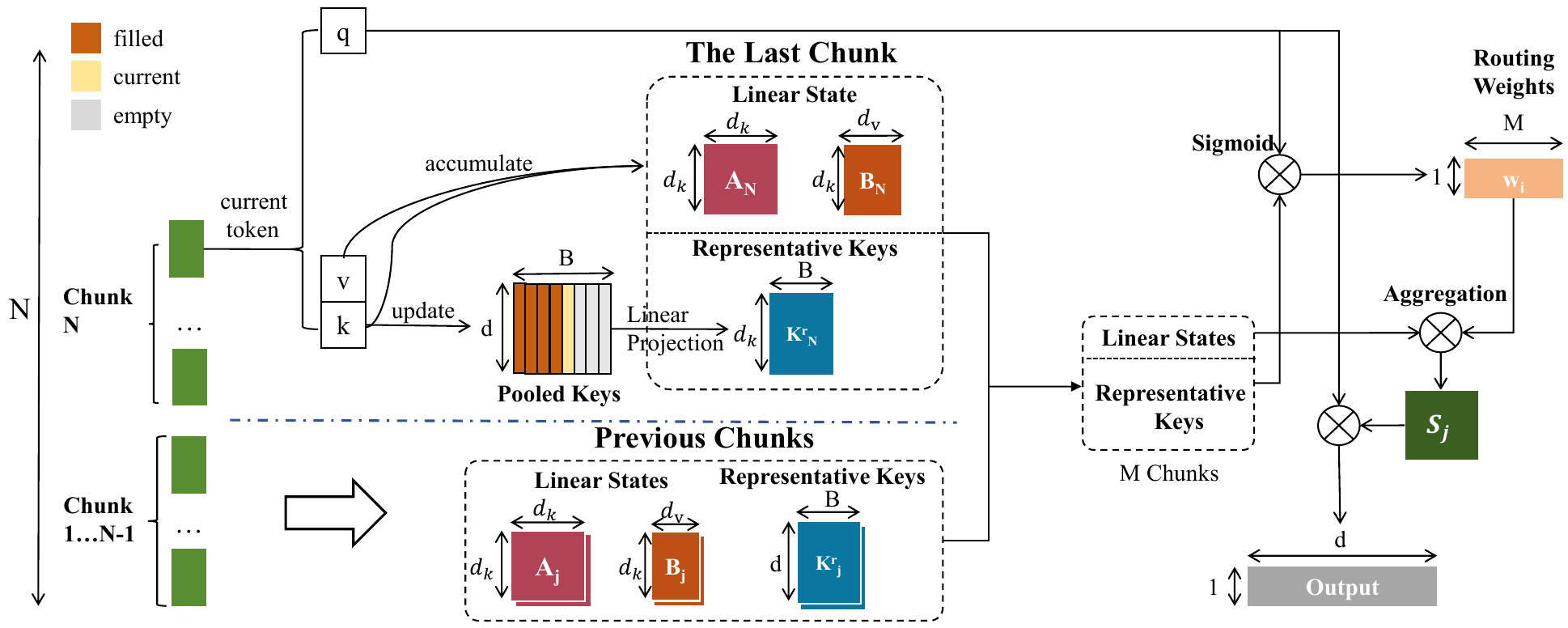}
    \caption{
\textbf{Overview of HLA.}
Each completed chunk is represented by an affine transition and a set of
self-attentively pooled routing representatives. For each query token,
independent sigmoid gates interpolate historical transitions with the
identity map. The gated transitions are composed in chronological order,
followed by the causal prefix transition of the current chunk and the
native GDN readout. The current chunk always has weight one.
}
    \label{fig:method-overview}
\end{figure}

\section{Methodology}

We first express its recurrence as chunk-wise affine transitions and then introduce query-dependent chunk-level attention to modulate their contributions to the recurrent state. Finally, we describe the training objective that encourages selective history access for sparse inference.

\subsection{Gated DeltaNet as Chunk-Wise Affine Transitions}
\label{sec:gdn-affine}

Based on the above observations, we instantiate our design on GDN~\citep{yanggated}, a modern recurrent linear-attention architecture. We
first express its recurrence in chunk-wise form and then introduce
query-dependent attention over historical chunks.

\noindent\textbf{Token-level GDN transition.}
We describe the computation for a single layer and head, using row-vector
queries, keys, and values. Let $q_t,k_t\in\mathbb{R}^{1\times d_k}$ denote
the projected query and key before the short convolution, and let
$\bar{q}_t,\bar{k}_t$ denote their convolved, $\ell$2-normalized counterparts.
Let $v_t\in\mathbb{R}^{1\times d_v}$ be the value used in the GDN update,
with scalar update strength $\beta_t$ and log-decay $g_t$. The recurrent
state $S_t\in\mathbb{R}^{d_k\times d_v}$ evolves as
\begin{equation}
S_t = A_t S_{t-1} + B_t,
\label{eq:gdn-affine-token}
\end{equation}
where
\begin{equation}
A_t =
\exp(g_t)\left(I-\beta_t\bar{k}_t^\top\bar{k}_t\right),
\qquad
B_t = \beta_t\bar{k}_t^\top v_t.
\label{eq:gdn-ab}
\end{equation}
Here, $A_t\in\mathbb{R}^{d_k\times d_k}$,
$B_t\in\mathbb{R}^{d_k\times d_v}$, and $I$ is the $d_k$-dimensional
identity matrix.

\noindent\textbf{Chunk-wise transition.}
We partition a sequence of length $T$ into $N=\lceil T/L\rceil$ contiguous
chunks of at most $L$ tokens, denoted by
$\{\mathcal{C}_j\}_{j=1}^{N}$. Let $c(i)$ denote the index of the chunk
containing query token $i$. Since affine maps are closed under composition,
the token-level updates within chunk $j$ can be summarized exactly by
$(A_j,B_j)$, with the same dimensions as their token-level counterparts.
At chunk granularity, let $S_{j-1}$ and $S_j$ denote the states immediately
before and after chunk $j$, respectively. Then
\begin{equation}
S_j = A_j S_{j-1} + B_j.
\label{eq:chunk-affine}
\end{equation}
The pair $(A_j,B_j)$ captures the chunk's complete effect on any incoming
state: $B_j$ represents its additive contribution, while $A_j$ determines
how it transforms earlier memory.

\subsection{Query-Dependent Chunk-Level Attention}
\label{sec:chunk-attention}

We estimate chunk relevance from
a small set of representative keys. Unlike softmax attention, which mixes
chunk outputs with normalized weights, GDN composes sequential state
transitions. We therefore use independent sigmoid attention weights,
allowing historical transitions to be modulated without a sum-to-one
constraint.

\noindent\textbf{Pooled chunk representations.}
Each completed chunk is divided into $P=L/p$ non-overlapping pooling
windows of $p$ tokens. For window $b$ of chunk $j$, let
$X_{j,b}\in\mathbb{R}^{p\times d_k}$ stack its pre-convolution keys.
We introduce head-wise learnable projections
$W_Q^r,W_K^p,W_V^p,W_O^p\in\mathbb{R}^{d_k\times d_k}$ and compute
\begin{equation}
q_i^r=q_iW_Q^r,
\qquad
K_{j,b}^p=X_{j,b}W_K^p,
\qquad
V_{j,b}^p=X_{j,b}W_V^p.
\label{eq:routing-query-proj}
\end{equation}
The representative key is obtained by self-attention within the window,
followed by mean pooling and an output projection:
\begin{equation}
\bar{k}_{j,b}^r
=
\operatorname{Mean}_{\mathrm{rows}}
\left[
\operatorname{softmax}
\left(
\frac{K_{j,b}^p(K_{j,b}^p)^\top}{\sqrt{d_k}}
\right)
V_{j,b}^p
\right]W_O^p.
\label{eq:self-attention-pool}
\end{equation}
Although pooling is bidirectional within each window, its representatives
are used only after the corresponding chunk has been completed. Queries
within the current chunk rely only on its causal prefix transition. Here, softmax is applied row-wise, and
$\operatorname{Mean}_{\mathrm{rows}}$ averages the token outputs within
the window. This produces one representative
$\bar{k}_{j,b}^r\in\mathbb{R}^{1\times d_k}$ per window.

\noindent\textbf{Chunk-level attention weights.}
For each historical chunk $j<c(i)$, we aggregate query--representative
similarities using log-mean-exp:
\begin{equation}
s_{i,j}
=
\log
\left[
\frac{1}{P}
\sum_{b=1}^{P}
\exp
\left(
\frac{q_i^r(\bar{k}_{j,b}^r)^\top}{\sqrt{d_k}}
\right)
\right].
\label{eq:chunk-score}
\end{equation}
The chunk-level attention weight is
\begin{equation}
w_{i,j}=\sigma(s_{i,j}),
\label{eq:chunk-attn-weight}
\end{equation}
where $\sigma(x)=1/(1+\exp(-x))$ denotes the sigmoid function.
We set $w_{i,c(i)}=1$ for the current chunk and mask future chunks with
$w_{i,j}=0$ for $j>c(i)$. During training, historical weights remain
continuous. At inference, weights below the threshold $\theta=0.1$ are
set to zero, while the remaining weights retain their sigmoid values
without renormalization.

\subsection{Query-Conditioned Historical States}
\label{sec:affine-history}

The chunk-level attention weights modulate complete GDN transitions
rather than merely scaling chunk readouts. For query token $i$, let
$S_{i,j}\in\mathbb{R}^{d_k\times d_v}$ denote its query-conditioned state
after processing historical chunks $1,\ldots,j$. Starting from
$S_{i,0}=0$, we apply each historical chunk $j<c(i)$ in chronological
order:
\begin{equation}
S_{i,j}
=
\left[(1-w_{i,j})I+w_{i,j}A_j\right]S_{i,j-1}
+
w_{i,j}B_j.
\label{eq:routed-affine}
\end{equation}
A weight of one applies the original chunk transition, whereas a weight
of zero leaves the incoming state unchanged. Intermediate weights
interpolate between these cases. Crucially, the attention weight $w_{i,j}$
modulates both the chunk's additive contribution and its transformation
of earlier memory.

The current chunk uses its original GDN updates because
$w_{i,c(i)}=1$. Let $(A_{c(i),\leq i},B_{c(i),\leq i})$ denote the affine
transition over its visible prefix, up to and including token $i$.
The final query-conditioned state is
\begin{equation}
S_i^{\methodname}
=
A_{c(i),\leq i}S_{i,c(i)-1}
+
B_{c(i),\leq i}.
\label{eq:hla-state}
\end{equation}
We apply the original GDN readout
$\bar{q}_iS_i^{\methodname}/\sqrt{d_k}$, followed by the unchanged gated
normalization and output projection. Setting all visible attention
weights to one recovers the original GDN recurrence; otherwise,
\methodname{} enables query-dependent history construction while reusing
the same chunk-level affine summaries.

\subsection{Training Objective}
\label{sec:training-objective}

To enable sparse inference, we encourage chunk-level attention to
concentrate on a small subset of historical chunks. For pretrained
adaptation, we freeze the backbone parameters and optimize the routing
and pooling modules with
\begin{equation}
\mathcal{L}
=
\mathcal{L}_{\mathrm{LM}}
+\lambda_b\mathcal{L}_{\mathrm{budget}}
+\lambda_s\mathcal{L}_{\mathrm{support}},
\label{eq:training-objective}
\end{equation}
where $\mathcal{L}_{\mathrm{LM}}$ is the next-token cross-entropy loss.

For query $i$, we define the total historical routing mass
$G_i=\sum_{j<c(i)}w_{i,j}$ and normalize the historical weights as
\begin{equation}
\pi_{i,j}
=
\frac{\max(w_{i,j},\epsilon)}
{\sum_{k<c(i)}\max(w_{i,k},\epsilon)}.
\end{equation}
Inspired by the order-2 effective number introduced by
\citet{diversity1973hill} and its recent use for quantifying effective
model components~\citep{wang2026effective}, we measure the effective support of the normalized historical routing
weights as
\begin{equation}
K_i^{\mathrm{eff}}
=
\left(\sum_{j<c(i)}\pi_{i,j}^{2}\right)^{-1}.
\label{eq:effective-support}
\end{equation}
This quantity equals $K$ for a uniform distribution over $K$ chunks and
decreases as the routing mass becomes more concentrated.

We then use
\begin{equation}
\begin{aligned}
\mathcal{L}_{\mathrm{budget}}
&=
\mathbb{E}\left[
\max\left(0,\frac{G_i-K_b}{K_b}\right)^2
\right],
\mathcal{L}_{\mathrm{support}}
=
\mathbb{E}\left[
\max\left(0,K_i^{\mathrm{eff}}-K_s\right)^2
\right].
\end{aligned}
\label{eq:attention-regularization}
\end{equation}
The budget term penalizes excessive historical routing mass, while the
support term encourages concentrated routing without imposing a hard
bound on the number of accessed chunks. Both regularizers exclude the
current chunk.

Training uses continuous routing weights. At inference, weights below
$\theta$ are set to zero without renormalization, introducing an
approximation to the continuous routing computation. Once a gate is set
to zero, skipping the corresponding identity transition is exact for the
thresholded model, reducing affine-state memory traffic and computation.

\section{Experiments}

\subsection{Implementation Details}

\paragraph{Training setup.}
We evaluate HLA under pretrained adaptation and from-scratch training.
For pretrained adaptation, we use Qwen3.5-Base models at the 0.8B, 2B,
4B, and 9B scales, freezing the backbone and optimizing only the routing
and pooling modules in GDN layers. Training uses 4096-token sequences,
$L=256$, and $p=16$. We set $K_b=3$, $K_s=3$,
$\lambda_b=0.005$, and $\lambda_s=0.001$. The router is optimized with
AdamW using a peak learning rate of $10^{-4}$, linear warm-up, and
cosine decay.
For the from-scratch setting, we train 1.3B-scale HLA and GDN models on
SlimPajama for 100B tokens with a 4K context under the same global batch
and optimization budget. HLA uses $L=256$ and $p=16$. All model
parameters are trained without routing regularization.

\paragraph{Evaluation.}
For pretrained Qwen3.5 models, evaluation uses $L=256$, $p=32$, and
$\theta=0.1$. We evaluate LongBench-V2 under a 4K context budget and all
13 RULER tasks at 4K, comparing HLA with the corresponding Native GDN
and MHLA baselines. For the from-scratch models, we evaluate the final
checkpoints on RULER at 4K, 8K, 16K, and 32K.

\subsection{Main Results}

\begin{table*}[t]
    \centering
    \caption{
    \textbf{From-scratch 1.3B GDN results on RULER from 4K to 32K.}
    HLA and GDN are trained from scratch with the same 100B-token budget
    and a 4K training context. We report the macro average and the full
    task-level breakdown using 500 matched examples per task and context
    length. The better result within each context length is highlighted
    in bold.
    }
    \label{tab:ruler-extrapolation}
    \scriptsize
    \setlength{\tabcolsep}{2.6pt}
    \resizebox{\textwidth}{!}{%
    \begin{tabular}{clc|ccc|ccc|cc|c|cc|cc}
        \toprule
        \multirow{2}{*}{Context}
        & \multirow{2}{*}{Method}
        & \multirow{2}{*}{Avg.}
        & \multicolumn{3}{c|}{Single Needle}
        & \multicolumn{3}{c|}{Multi-Key}
        & \multicolumn{2}{c|}{Multi-Value/Query}
        & \multirow{2}{*}{VT}
        & \multicolumn{2}{c|}{Word Extraction}
        & \multicolumn{2}{c}{QA} \\
        \cmidrule(lr){4-6}
        \cmidrule(lr){7-9}
        \cmidrule(lr){10-11}
        \cmidrule(lr){13-14}
        \cmidrule(lr){15-16}
        & & &
        S1 & S2 & S3 &
        MK1 & MK2 & MK3 &
        MV & MQ &
        & CWE & FWE &
        SQuAD & Hotpot \\
        \midrule

        4K
        & GDN
        & 24.63
        & 99.00 & 66.00 & \textbf{46.00}
        & 26.60 & 0.00 & \textbf{0.00}
        & 15.80 & 15.60
        & \textbf{2.56}
        & \textbf{24.06} & \textbf{5.20}
        & 11.00 & 8.40 \\

        \rowcolor{gray!15}
        4K
        & \methodname{} (ours)
        & \textbf{25.46}
        & \textbf{100.00} & \textbf{68.00} & 31.20
        & \textbf{30.40} & \textbf{1.00} & \textbf{0.00}
        & \textbf{25.20} & \textbf{20.40}
        & 2.08
        & 18.24 & 0.07
        & \textbf{14.40} & \textbf{20.00} \\
        \midrule

        8K
        & GDN
        & 15.02
        & 56.60 & 29.80 & \textbf{20.60}
        & 20.40 & \textbf{0.00} & \textbf{0.00}
        & 8.50 & 7.65
        & \textbf{13.76}
        & \textbf{17.28} & \textbf{5.33}
        & 6.60 & 8.80 \\

        \rowcolor{gray!15}
        8K
        & \methodname{} (ours)
        & \textbf{17.69}
        & \textbf{100.00} & \textbf{31.80} & 15.80
        & \textbf{21.00} & \textbf{0.00} & \textbf{0.00}
        & \textbf{17.90} & \textbf{16.55}
        & 1.28
        & 2.72 & 0.13
        & \textbf{7.80} & \textbf{15.00} \\
        \midrule

        16K
        & GDN
        & 7.98
        & 23.80 & \textbf{9.40} & 2.40
        & \textbf{9.60} & \textbf{0.00} & \textbf{0.00}
        & 3.90 & 1.90
        & \textbf{22.64}
        & \textbf{6.08} & \textbf{9.80}
        & 5.20 & 9.00 \\

        \rowcolor{gray!15}
        16K
        & \methodname{} (ours)
        & \textbf{11.05}
        & \textbf{97.00} & 5.00 & \textbf{3.60}
        & 4.60 & \textbf{0.00} & \textbf{0.00}
        & \textbf{6.00} & \textbf{4.50}
        & 1.80
        & 2.94 & 0.20
        & \textbf{7.00} & \textbf{11.00} \\
        \midrule

        32K
        & GDN
        & 3.65
        & 11.20 & \textbf{2.40} & 2.40
        & \textbf{5.60} & \textbf{0.00} & \textbf{0.00}
        & 0.50 & 0.15
        & 3.40
        & 1.54 & \textbf{5.67}
        & \textbf{5.40} & 9.20 \\

        \rowcolor{gray!15}
        32K
        & \methodname{} (ours)
        & \textbf{7.87}
        & \textbf{67.40} & 0.80 & \textbf{3.00}
        & 2.40 & \textbf{0.00} & \textbf{0.00}
        & \textbf{1.90} & \textbf{2.00}
        & \textbf{4.92}
        & \textbf{4.04} & 0.27
        & 4.80 & \textbf{10.80} \\
        \bottomrule
    \end{tabular}}
\end{table*}

\begin{table*}[t]
    \centering
    \caption{
    \textbf{RULER results across Qwen3.5 model scales.}
    We report the macro average over all 13 tasks together with the complete
    task-level breakdown. Each task contains 500 examples. The best result
    within each model scale is highlighted in bold.
    }
    \label{tab:ruler-results}
    \scriptsize
    \setlength{\tabcolsep}{2.5pt}
    \resizebox{\textwidth}{!}{%
    \begin{tabular}{clc|ccc|ccc|cc|c|cc|cc}
        \toprule
        \multirow{2}{*}{Scale}
        & \multirow{2}{*}{Method}
        & \multirow{2}{*}{Avg.}
        & \multicolumn{3}{c|}{Single Needle}
        & \multicolumn{3}{c|}{Multi-Key}
        & \multicolumn{2}{c|}{Multi-Value/Query}
        & \multirow{2}{*}{VT}
        & \multicolumn{2}{c|}{Word Extraction}
        & \multicolumn{2}{c}{QA} \\
        \cmidrule(lr){4-6}
        \cmidrule(lr){7-9}
        \cmidrule(lr){10-11}
        \cmidrule(lr){13-14}
        \cmidrule(lr){15-16}
        & & & S1 & S2 & S3
        & MK1 & MK2 & MK3
        & MV & MQ
        & & CWE & FWE
        & SQuAD & Hotpot \\
        \midrule

        \multirow{3}{*}{0.8B}
        & Native GDN
        & 79.141
        & \textbf{100.00} & 94.83 & \textbf{96.00}
        & 95.00 & 96.00 & 85.00
        & 95.00 & \textbf{90.00}
        & 74.00 & 55.00 & 70.00
        & 42.00 & 36.00 \\

        & MHLA
        & 80.182
        & \textbf{100.00} & 95.33 & 95.60
        & 95.30 & 96.50 & 86.00
        & 95.80 & 89.70
        & 76.20 & 58.20 & 71.50
        & 44.50 & 37.74 \\

        \rowcolor{gray!15}
        & \methodname{} (ours)
        & \textbf{83.115}
        & \textbf{100.00} & \textbf{96.83} & 95.00
        & \textbf{96.00} & \textbf{98.00} & \textbf{90.00}
        & \textbf{97.00} & 89.00
        & \textbf{81.00} & \textbf{67.00} & \textbf{76.00}
        & \textbf{52.00} & \textbf{42.67} \\
        \midrule

        \multirow{3}{*}{2B}
        & Native GDN
        & 92.092
        & \textbf{100.00} & \textbf{100.00} & \textbf{100.00}
        & 99.20 & \textbf{100.00} & 99.80
        & 99.70 & \textbf{98.55}
        & 94.00 & 90.90 & 90.40
        & 69.25 & 55.40 \\

        & MHLA
        & 92.192
        & \textbf{100.00} & \textbf{100.00} & 99.80
        & 99.50 & \textbf{100.00} & \textbf{100.00}
        & 99.85 & 98.20
        & 94.30 & 91.20 & 90.70
        & 69.60 & 55.34 \\

        \rowcolor{gray!15}
        & \methodname{} (ours)
        & \textbf{93.342}
        & \textbf{100.00} & \textbf{100.00} & 99.80
        & \textbf{99.60} & \textbf{100.00} & \textbf{100.00}
        & \textbf{99.90} & 98.25
        & \textbf{95.50} & \textbf{92.90} & \textbf{91.60}
        & \textbf{74.25} & \textbf{61.65} \\
        \midrule

        \multirow{3}{*}{4B}
        & Native GDN
        & 87.871
        & \textbf{100.00} & \textbf{100.00} & \textbf{100.00}
        & 99.40 & \textbf{100.00} & \textbf{100.00}
        & 99.95 & \textbf{100.00}
        & \textbf{100.00} & 87.74 & 99.00
        & 26.03 & 30.20 \\

        & MHLA
        & 88.050
        & \textbf{100.00} & \textbf{100.00} & \textbf{100.00}
        & \textbf{99.60} & \textbf{100.00} & 99.90
        & \textbf{100.00} & \textbf{100.00}
        & \textbf{100.00} & 88.50 & 99.20
        & 27.50 & 29.95 \\

        \rowcolor{gray!15}
        & \methodname{} (ours)
        & \textbf{89.272}
        & \textbf{100.00} & 99.80 & \textbf{100.00}
        & \textbf{99.60} & \textbf{100.00} & \textbf{100.00}
        & 99.65 & \textbf{100.00}
        & \textbf{100.00} & \textbf{90.24} & \textbf{99.50}
        & \textbf{34.03} & \textbf{37.71} \\
        \midrule

        \multirow{3}{*}{9B}
        & Native GDN
        & 90.115
        & \textbf{100.00} & \textbf{100.00} & \textbf{100.00}
        & 99.80 & \textbf{100.00} & \textbf{100.00}
        & 99.90 & \textbf{100.00}
        & \textbf{100.00} & 93.26 & 98.80
        & 37.13 & 42.60 \\

        & MHLA
        & 90.200
        & \textbf{100.00} & \textbf{100.00} & \textbf{100.00}
        & \textbf{99.90} & \textbf{100.00} & \textbf{100.00}
        & \textbf{100.00} & \textbf{100.00}
        & \textbf{100.00} & 93.00 & 99.00
        & 37.50 & 43.20 \\

        \rowcolor{gray!15}
        & \methodname{} (ours)
        & \textbf{91.128}
        & \textbf{100.00} & \textbf{100.00} & 99.80
        & \textbf{99.90} & \textbf{100.00} & \textbf{100.00}
        & 99.70 & \textbf{100.00}
        & \textbf{100.00} & \textbf{95.26} & \textbf{99.20}
        & \textbf{44.13} & \textbf{46.67} \\
        \bottomrule
    \end{tabular}}
\end{table*}

\begin{table*}[t]
    \centering
    \caption{
    \textbf{LongBench-V2 results across Qwen3.5 model scales.}
    We report overall accuracy (\%) together with breakdowns by difficulty
    and original context-length category. The best result within each model
    scale is highlighted in bold.
    }
    \label{tab:longbenchv2-results}
    \scriptsize
    \setlength{\tabcolsep}{5.0pt}
    \resizebox{0.75\textwidth}{!}{%
    \begin{tabular}{clc|cc|ccc}
        \toprule
        \multirow{2}{*}{Scale}
        & \multirow{2}{*}{Method}
        & \multirow{2}{*}{Overall}
        & \multicolumn{2}{c|}{Difficulty}
        & \multicolumn{3}{c}{Context Length} \\
        \cmidrule(lr){4-5}
        \cmidrule(lr){6-8}
        & & & Easy & Hard & Short & Medium & Long \\
        \midrule

        \multirow{3}{*}{0.8B}
        & Native GDN
        & 22.86 & 23.96 & 22.19 & 25.56 & 19.07 & 25.93 \\
        & MHLA
        & 23.26 & 23.96 & 22.83 & \textbf{26.11} & 19.53 & 25.93 \\
        \rowcolor{gray!15}
        & \methodname{} (ours)
        & \textbf{28.43} & \textbf{27.60} & \textbf{28.94}
        & 25.00 & \textbf{27.44} & \textbf{36.11} \\
        \midrule

        \multirow{3}{*}{2B}
        & Native GDN
        & 27.83 & 31.25 & 25.72 & 26.67 & 29.77 & 25.93 \\
        & MHLA
        & 27.44 & \textbf{31.77} & 24.76 & 26.67 & 29.77 & 24.07 \\
        \rowcolor{gray!15}
        & \methodname{} (ours)
        & \textbf{29.82} & 30.73 & \textbf{29.26}
        & \textbf{26.67} & \textbf{33.02} & \textbf{28.70} \\
        \midrule

        \multirow{3}{*}{4B}
        & Native GDN
        & 29.62 & 31.77 & \textbf{28.30} & \textbf{33.89} & 28.84 & 24.07 \\
        & MHLA
        & 30.02 & 32.81 & \textbf{28.30} & \textbf{33.89} & 30.23 & 23.15 \\
        \rowcolor{gray!15}
        & \methodname{} (ours)
        & \textbf{31.01} & \textbf{35.94} & 27.97
        & 28.33 & \textbf{34.42} & \textbf{28.70} \\
        \midrule

        \multirow{3}{*}{9B}
        & Native GDN
        & 33.00 & 34.38 & 32.15 & 33.89 & 33.49 & 30.56 \\
        & MHLA
        & 33.00 & \textbf{34.90} & 31.83 & \textbf{34.44} & 32.56 & 31.48 \\
        \rowcolor{gray!15}
        & \methodname{} (ours)
        & \textbf{34.19} & 34.89 & \textbf{33.76}
        & 31.67 & \textbf{35.81} & \textbf{35.19} \\
        \bottomrule
    \end{tabular}}
\end{table*}

\textbf{From-scratch 1.3B GDN results on RULER.}~
We evaluate HLA in a from-scratch setting to examine
whether its advantage persists beyond the training context. HLA and GDN
are trained with the same 100B-token budget and a 4K training context,
and are evaluated on the full RULER suite from 4K to 32K. This experiment
complements the Qwen3.5 adaptation results by removing the effect of
pretrained initialization and directly testing long-context
generalization under matched training conditions.
As shown in Table~\ref{tab:ruler-extrapolation}, HLA achieves higher
macro-average scores at every evaluated context length, improving GDN by
0.83, 2.67, 3.07, and 4.22 points at 4K, 8K, 16K, and 32K,
respectively. The performance gap becomes more pronounced beyond the 4K
training context, indicating that HLA retains a stronger aggregate
capability as the sequence length increases.
The improvements are task-dependent rather than uniform. HLA shows
particularly large gains on long-range retrieval, with S1 improving from
56.6 to 100.0 at 8K, from 23.8 to 97.0 at 16K, and from 11.2 to 67.4
at 32K. It also consistently improves MV and MQ across all evaluated
context lengths, together with gains on several QA settings. At the same
time, performance decreases on some VT, CWE, and FWE cases. Overall,
the results suggest that query-dependent access to chunk-level recurrent
memory is especially beneficial for preserving and retrieving sparse
historical information over long trajectories, while its effect remains
task-dependent for other forms of long-context computation.

\textbf{Qwen3.5 long-context evaluation.}~
We first evaluate \methodname{} by adapting pretrained Qwen3.5 models
from 0.8B to 9B parameters. On LongBench-V2~\citep{bai2025longbenchv2},
Table~\ref{tab:longbenchv2-results} shows that \methodname{} improves
Native GDN at all four scales, with overall gains of 5.57, 1.99, 1.39,
and 1.19 percentage points, respectively. The largest improvement occurs
at 0.8B, where \methodname{} also substantially outperforms MHLA
(28.43\% vs.\ 23.26\%). The gains are particularly pronounced on more
challenging and longer examples: at 0.8B, accuracy improves from 22.19\%
to 28.94\% on the Hard subset, from 19.07\% to 27.44\% on Medium
contexts, and from 25.93\% to 36.11\% on Long contexts.
On RULER~\citep{hsieh2024ruler}, Table~\ref{tab:ruler-results} further
shows that \methodname{} achieves the highest average score at every
model scale, outperforming Native GDN by 3.974, 1.250, 1.401, and
1.013 points and MHLA by 2.933, 1.150, 1.222, and 0.928 points from
0.8B to 9B. The improvements are especially clear on aggregation and
question-answering tasks; for example, at 0.8B, \methodname{} improves
CWE from 55.00 to 67.00, SQuAD from 42.00 to 52.00, and HotpotQA from
36.00 to 42.67 over Native GDN. Together, these results show that
query-dependent historical aggregation improves long-context modeling
across model scales and complementary evaluation settings.

\subsection{Ablation Study}
\begin{figure*}[t]
    \centering

    \begin{subfigure}[t]{0.32\textwidth}
        \centering
        \includegraphics[width=\linewidth]
        {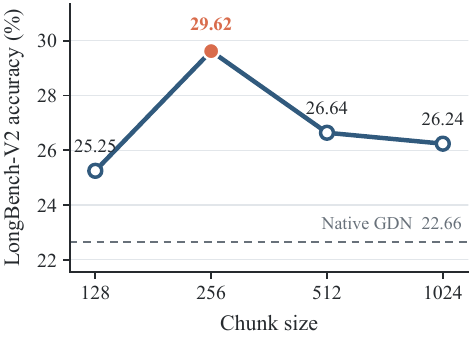}
        \caption{Effect of chunk size.}
        \label{fig:chunk_size}
    \end{subfigure}
    \hfill
    \begin{subfigure}[t]{0.32\textwidth}
        \centering
        \includegraphics[width=\linewidth]
        {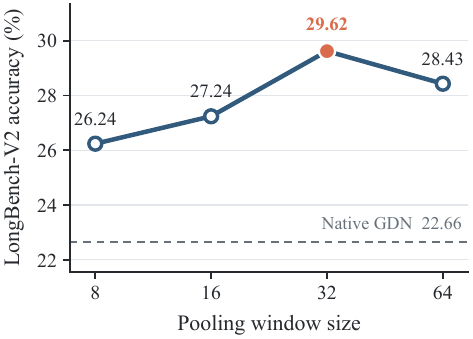}
        \caption{Effect of pooling window size.}
        \label{fig:pooling_size}
    \end{subfigure}
    \hfill
    \begin{subfigure}[t]{0.32\textwidth}
        \centering
        \includegraphics[width=\linewidth]
        {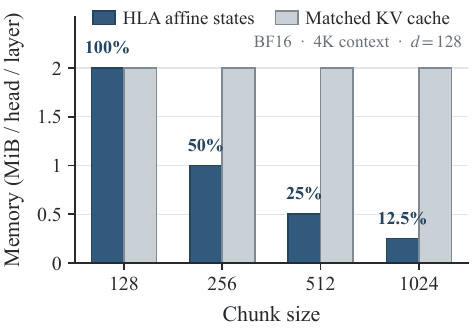}
        \caption{Memory scaling with chunk size.}
        \label{fig:memory_scaling}
    \end{subfigure}

    \caption{
    Ablation analyses of chunk size, pooling window size, and historical-memory scaling.
    }
    \label{fig:ablation_chunk_pool_memory}
\end{figure*}

\noindent\textbf{Effect of chunk size.}
We study the effect of chunk size on long-context performance. As shown in Figure~\ref{fig:chunk_size}, increasing the chunk size from $128$ to $256$ substantially improves LongBench-V2 accuracy from $25.25\%$ to $29.62\%$. However, further increasing the chunk size leads to a clear performance drop, with accuracies of $26.64\%$ and $26.24\%$ for chunk sizes $512$ and $1024$, respectively. This non-monotonic trend is consistent with a trade-off between retrieval
granularity and contextual completeness. Smaller chunks provide finer-grained retrieval but divide the context into fragmented local units, whereas larger chunks aggregate more information into each memory unit but reduce the flexibility of selectively accessing relevant content. An intermediate chunk size therefore provides a better balance between contextual completeness and retrieval granularity. We adopt a chunk size of $256$, which achieves the best LongBench-V2 performance among the evaluated settings.

\noindent\textbf{Effect of pooling window size.}
We study the pooling window used to construct chunk-level routing representations. As shown in Figure~\ref{fig:pooling_size}, LongBench-V2 accuracy improves from $26.24\%$ to $27.24\%$ and $29.62\%$ as the pooling window increases from $8$ to $16$ and $32$, respectively. Increasing the window further to $64$ reduces the accuracy to $28.43\%$. These results suggest that the routing representation benefits from moderate local aggregation: a small pooling window may retain excessive token-level redundancy and produce less stable chunk summaries, whereas an overly large window can discard fine-grained information useful for distinguishing relevant historical chunks. A pooling window of $32$ provides the best balance between compactness and discriminative routing information and is therefore used as the default setting.

\noindent\textbf{Memory scaling with chunk size.}
We analyze the historical-memory cost of the proposed chunk-wise affine states under a fixed $4$K context length and head dimension $d=128$. As shown in Figure~\ref{fig:memory_scaling}, the matched KV cache requires $2$ MiB per layer and head and remains unchanged with the chunk size, since it stores token-level keys and values for the entire context. In contrast, the memory required by the affine states decreases approximately inversely with the chunk size because a larger chunk size produces fewer chunk-level summaries. Specifically, the affine-state memory is equal to the matched KV-cache cost at a chunk size of $128$, but decreases to $50\%$, $25\%$, and $12.5\%$ of the KV-cache memory at chunk sizes $256$, $512$, and $1024$, respectively. Notably, the default chunk size of $256$ already reduces historical-state storage by $2\times$ while achieving the highest LongBench-V2 accuracy. This demonstrates that chunk-wise state representation provides a favorable trade-off between addressable long-range memory, retrieval granularity, and storage efficiency.

\begin{table}[t]
    \centering
    \caption{
    Prefill and decoding latency of Native GDN and \methodname{} across
    Qwen3.5 model scales. We use a single NVIDIA H200 NVL, BF16,
    batch size 1, a 4096-token prompt, and 128-token steady-state decoding.
    \methodname{} uses the same inference configuration as the main
    experiments.
    }
    \label{tab:efficiency}
    \small
    \begin{tabular}{lccc|ccc}
        \toprule
        & \multicolumn{3}{c|}{Prefill} & \multicolumn{3}{c}{Decode} \\
        \cmidrule(lr){2-4} \cmidrule(lr){5-7}
        Scale
        & GDN (ms)
        & \methodname{} (ms)
        & Ratio
        & GDN (ms/tok)
        & \methodname{} (ms/tok)
        & Ratio \\
        \midrule
        0.8B
        & 40.36 & 46.00 & 1.140$\times$
        & 5.90 & 7.01 & 1.188$\times$ \\
        2B
        & 53.89 & 59.17 & 1.098$\times$
        & 6.49 & 7.60 & 1.171$\times$ \\
        4B
        & 115.09 & 132.29 & 1.149$\times$
        & 10.30 & 11.96 & 1.161$\times$ \\
        9B
        & 159.21 & 178.97 & 1.124$\times$
        & 12.19 & 13.88 & 1.139$\times$ \\
        \bottomrule
    \end{tabular}
\end{table}

\paragraph{Inference efficiency.}
Table~\ref{tab:efficiency} compares the inference latency of
\methodname{} with the corresponding Native GDN backbones under the same
operating point as our main accuracy evaluation. Across model scales,
\methodname{} incurs a moderate prefill overhead of 9.8--14.9\% and a
steady-state decoding overhead of 13.9--18.8\%. The relative decoding
overhead generally decreases as the model scale increases, from
1.188$\times$ at 0.8B to 1.139$\times$ at 9B, suggesting that the
additional chunk-level routing cost becomes less significant relative
to the backbone computation at larger scales. Overall, \methodname{}
improves long-context selectivity while introducing only a modest
latency overhead over the original recurrent GDN inference path.


\section{Conclusion}

We present HLA, a query-dependent chunk-level attention framework for
GDN. HLA represents completed chunks as affine state transitions and
dynamically modulates their composition according to the current query,
controlling both new memory and the transformation of earlier history.
Across pretrained Qwen3.5 models, HLA improves long-context performance
over native GDN and fixed chunk mixing. Controlled from-scratch
experiments further show consistent aggregate gains on RULER from 4K to
32K, including contexts substantially longer than those seen during
training. These results demonstrate the value of query-dependent
composition of compact recurrent-memory summaries for long-context
modeling.

\noindent\textbf{Limitations.}
Our current evaluation focuses on standard long-context language understanding benchmarks and does not yet cover more demanding interactive settings in which long-range memory is continuously accumulated and selectively reused. Agentic systems are a particularly important example: long-running agents must maintain goals, plans, observations, retrieved evidence, and tool feedback over extended trajectories, while repeatedly accessing only a small subset of this history for each decision. Such settings may place substantially stronger demands on adaptive historical retrieval than static document understanding. Evaluating \methodname{} in long-horizon agentic workloads, including multi-turn tool use and persistent memory, is therefore an important direction for future work.



\clearpage
\bibliographystyle{unsrtnat}
\bibliography{iclr2027_conference}

\clearpage
\appendix
\section{Appendix}


\subsection{Algorithmic Sketches}
\label{sec:algorithmic-sketches}

Algorithms~\ref{alg:appendix-mix-gdn-prefill} and
\ref{alg:appendix-mix-gdn-decode} describe the current GDN-based
\methodname{} for training/prefill and dynamic autoregressive decoding.
We use the notation of the main text: $q_i,k_i$ are pre-convolution
features, whereas $\bar{q}_i,\bar{k}_i$ are the convolved and normalized
features used by the native recurrence. All operations are shown for
one layer and value head, with row-vector queries and keys. The native
convolution is applied before chunk partitioning; its cache is preserved
across chunk boundaries during decoding. The algorithms return
pre-normalization readouts, followed by the original gated normalization
and output projection.

Let $L$ be the chunk size, $p$ the pooling-window size, and $P=L/p$ the
number of representatives per completed chunk, with $p$ dividing $L$.
The current-chunk index is $c(i)=1+\lfloor(i-1)/L\rfloor$. We write
$U_j\in\mathbb{R}^{P\times d_k}$ for the stacked representatives of
chunk $j$. The routines \textsc{Pool} and \textsc{ChunkAttention}
implement window-level self-attention pooling and log-mean-exp sigmoid
weighting, respectively. Only completed historical chunks are scored;
the current chunk contributes through its causal prefix with weight one.
Training uses continuous weights, while inference sets historical weights
below $\theta$ to zero without renormalization.

\noindent\textbf{Exact prefix data.}
For token $i$ in chunk $j$, \textsc{ChunkData} produces the final summary
$(A_j,B_j)$ and the following current-prefix quantities:
\begin{equation}
o_i^{\mathrm{loc}}
=
\frac{\bar{q}_i B_{j,\leq i}}{\sqrt{d_k}},
\qquad
r_i^{\mathrm{loc}}
=
\frac{\bar{q}_i A_{j,\leq i}}{\sqrt{d_k}},
\label{eq:appendix-prefix-readout}
\end{equation}
where $(A_{j,\leq i},B_{j,\leq i})$ summarizes the chunk prefix ending
at $i$, and $r_i^{\mathrm{loc}}$ is a row-vector readout of its incoming
state. These quantities can be computed jointly by extending the native
GDN values to $[v\mid 0]$ and initializing the state as $[0\mid I]$.
The resulting final state is $[B_j\mid A_j]$. The reference code below
uses an explicit recurrence for clarity.

\begin{algorithm}[h]
\caption{\methodname{} training and prefill}
\label{alg:appendix-mix-gdn-prefill}
\footnotesize
\begin{algorithmic}[1]
\Require Projected features $\{q_i,k_i,\bar{q}_i,\bar{k}_i,v_i,\beta_i,g_i\}_{i=1}^{T}$
\Require Chunk size $L$, pool window $p$, learned attention/pooling projections, inference threshold $\theta$
\Ensure Readouts $\{o_i\}_{i=1}^{T}$ and, at inference, a decode cache
\State Partition the sequence into $N=\lceil T/L\rceil$ chunks
\For{each chunk $j$ in parallel}
    \State $(A_j,B_j,\{o_i^{\mathrm{loc}},r_i^{\mathrm{loc}}\}_{i\in\mathcal{C}_j})\gets\textsc{ChunkData}(\mathcal{C}_j)$
    \If{$\mathcal{C}_j$ contains $L$ tokens}
        \State $U_j\gets\textsc{Pool}(\{k_i\}_{i\in\mathcal{C}_j},p)$
    \EndIf
\EndFor
\For{each query $i$ in parallel}
    \State $o_i\gets o_i^{\mathrm{loc}},\quad r_i\gets r_i^{\mathrm{loc}}$
    \State Compute $w_{i,j}\gets\textsc{ChunkAttention}(q_i,U_j)$ for $j<c(i)$
    \If{inference}
        \State Set historical $w_{i,j}<\theta$ to zero without renormalization
    \EndIf
\EndFor
\For{$j=N-1,\ldots,1$} \Comment{Newest to oldest}
    \For{each query $i$ with $c(i)>j$ in parallel}
        \If{training or $w_{i,j}\neq0$}
            \State $o_i\gets o_i+w_{i,j}r_iB_j$
            \State $r_i\gets(1-w_{i,j})r_i+w_{i,j}r_iA_j$
        \EndIf
    \EndFor
\EndFor
\If{inference}
    \State Cache completed $(A_j,B_j,U_j)$ and the active chunk's exact prefix and pooling state
\EndIf
\State \Return $\{o_i\}_{i=1}^{T}$ and the inference cache, if requested
\end{algorithmic}
\end{algorithm}

\noindent\textbf{Dynamic decode cache.}
The cache stores completed chunk summaries and representatives
$\{(A_j,B_j,U_j)\}_{j=1}^{M}$, the active chunk's affine pair
$(A_{\mathrm{cur}},B_{\mathrm{cur}})$, its completed pooling windows,
and the raw keys of its incomplete window. An empty active chunk starts
with $(I,0)$ and token count $n_{\mathrm{cur}}=0$.

\begin{algorithm}[t]
\caption{\methodname{} dynamic autoregressive decoding}
\label{alg:appendix-mix-gdn-decode}
\footnotesize
\begin{algorithmic}[1]
\Require New-token features $(q_i,k_i,\bar{q}_i,\bar{k}_i,v_i,\beta_i,g_i)$ and decode cache
\Require Chunk size $L$, pool window $p$, learned attention/pooling projections, threshold $\theta$
\Ensure Readout $o_i$ and updated cache
\State Form $(A_i,B_i)$ using Eq.~\ref{eq:gdn-ab}
\State $A_{\mathrm{cur}}\gets A_iA_{\mathrm{cur}},\quad B_{\mathrm{cur}}\gets A_iB_{\mathrm{cur}}+B_i$
\State Append $k_i$ to the active pooling window; increment $n_{\mathrm{cur}}$
\If{the pooling window contains $p$ tokens}
    \State Apply \textsc{Pool} to finalize its representative and clear the window buffer
\EndIf
\For{each completed chunk $j=1,\ldots,M$}
    \State $w_{i,j}\gets\textsc{ChunkAttention}(q_i,U_j)$
    \If{$w_{i,j}<\theta$}
        \State $w_{i,j}\gets0$
    \EndIf
\EndFor
\State $r\gets\bar{q}_i/\sqrt{d_k}$
\State $o_i\gets rB_{\mathrm{cur}},\quad r\gets rA_{\mathrm{cur}}$ \Comment{Current weight is one}
\For{$j=M,\ldots,1$} \Comment{Newest to oldest}
    \If{$w_{i,j}\neq0$}
        \State $o_i\gets o_i+w_{i,j}rB_j$
        \State $r\gets(1-w_{i,j})r+w_{i,j}rA_j$
    \EndIf
\EndFor
\If{$n_{\mathrm{cur}}=L$}
    \State Append the active $(A,B,U)$ to the completed-chunk cache
    \State Reset the active chunk to $(I,0)$ and clear its pooling state and token count
\EndIf
\State \Return $o_i$ and the updated cache
\end{algorithmic}
\end{algorithm}

The historical scan must proceed from newest to oldest, updating the
output before the readout vector. This evaluates the forward
query-conditioned recurrence without materializing a state matrix for
every query. Zero-weight transitions can be skipped, but their cached
summaries are retained because later queries may assign different weights.
Triton fusion and CUDA graph execution preserve these forward equations.

\section{PyTorch-Style Pseudocode}
\label{sec:pytorch-pseudocode}

The following mathematical forward reference covers one sequence, GDN
layer, and value head. It is not the optimized training or inference
implementation. Inputs $q,k$ contain pre-convolution features;
\texttt{qbar,kbar} contain the native convolved, normalized features.
Sequence inputs have shapes \texttt{[T,Dk]} for queries/keys,
\texttt{[T,Dv]} for values, and \texttt{[T]} for \texttt{beta,g}.
The enclosing model supplies native projection, convolution, head
repetition, gated normalization, and output projection. Python indices
are zero-based. The tuple \texttt{W} contains
$(W_Q^r,W_K^p,W_V^p,W_O^p)$, each of shape $d_k\times d_k$.
A common floating-point dtype is assumed below; production code uses
the required mixed-precision accumulation.

\subsection{Affine Updates and Prefix Readouts}

\begin{lstlisting}[style=pseudocode,breaklines=true]
import math
import torch


def affine_step(A, B, kbar, v, beta, g):
    # Exact rank-one token update: avoids materializing A_t.
    decay = g.exp()
    A_next = decay * (A - beta * torch.outer(kbar, kbar @ A))
    B_next = decay * (B - beta * torch.outer(kbar, kbar @ B))
    B_next = B_next + beta * torch.outer(kbar, v)
    return A_next, B_next


def chunk_data(qbar, kbar, v, beta, g):
    # Return final (A, B), local outputs, and prefix readouts.
    Dk, Dv = kbar.shape[-1], v.shape[-1]
    A = torch.eye(Dk, device=kbar.device, dtype=kbar.dtype)
    B = v.new_zeros(Dk, Dv)
    local, readout = [], []
    for t in range(len(kbar)):
        A, B = affine_step(A, B, kbar[t], v[t], beta[t], g[t])
        r = qbar[t] / math.sqrt(Dk)
        local.append(r @ B)
        readout.append(r @ A)
    return A, B, torch.stack(local), torch.stack(readout)
\end{lstlisting}

\subsection{Self-Attentive Pooling and Chunk Attention}

\begin{lstlisting}[style=pseudocode,breaklines=true]
def pool_keys(k, p, WK, WV, WO):
    # k: [tokens, Dk], containing complete pooling windows.
    if p <= 0 or len(k) % p:
        raise ValueError("pooling requires complete windows")
    Dk = k.shape[-1]
    if len(k) == 0:
        return k.new_empty(0, Dk)
    X = k.reshape(-1, p, Dk)
    Kp, Vp = X @ WK, X @ WV
    score = Kp @ Kp.transpose(-1, -2) / math.sqrt(Dk)
    return ((score.softmax(dim=-1) @ Vp) @ WO).mean(dim=1)


def chunk_weights(q, pooled, WQ, theta=None):
    # pooled contains completed historical chunks only.
    if len(pooled) == 0:
        return q.new_empty(0)
    U = torch.stack(pooled)                # [history, P, Dk]
    rq = q @ WQ
    score = torch.einsum("d,cpd->cp", rq, U) / math.sqrt(q.numel())
    score = score.logsumexp(dim=-1) - math.log(U.shape[1])
    w = score.sigmoid()
    if theta is not None:                 # inference only
        w = torch.where(w < theta, torch.zeros_like(w), w)
    return w                              # no renormalization
\end{lstlisting}

\subsection{Reverse Historical Readout}

\begin{lstlisting}[style=pseudocode,breaklines=true]
def reverse_history(out, r, A_hist, B_hist, w, skip_zeros=False):
    for j in range(len(A_hist) - 1, -1, -1):
        if skip_zeros and w[j].item() == 0:
            continue
        out = out + w[j] * (r @ B_hist[j])
        r = (1 - w[j]) * r + w[j] * (r @ A_hist[j])
    return out
\end{lstlisting}

\subsection{Training and Prefill Reference}

Set \texttt{theta=None} during training and use the inference threshold
for evaluation. The partial final chunk is processed at its true length;
fixed-shape kernels instead use neutral padding with $\beta=0$ and $g=0$.
The returned cache preserves only real tokens for continuation. The
per-query loop below is equivalent to the source-parallel schedule in
Algorithm~\ref{alg:appendix-mix-gdn-prefill}.

\begin{lstlisting}[style=pseudocode,breaklines=true]
def hla_prefill(q, k, qbar, kbar, v, beta, g, L, p, W, theta=None):
    # W = (WQ, WK_pool, WV_pool, WO_pool).
    # Returns pre-normalization readouts and an inference cache.
    if len(q) == 0 or L <= 0 or p <= 0 or L % p:
        raise ValueError("require T > 0 and L divisible by p")
    WQ, WK, WV, WO = W
    T, Dk = q.shape
    Dv = v.shape[-1]
    A, B, U, local, readout = [], [], [], [], []
    for start in range(0, T, L):
        end = min(start + L, T)
        sl = slice(start, end)
        a, b, o, r = chunk_data(qbar[sl], kbar[sl], v[sl], beta[sl], g[sl])
        A.append(a)
        B.append(b)
        local.append(o)
        readout.append(r)
        # A partial final chunk is never historical in this forward.
        U.append(pool_keys(k[sl], p, WK, WV, WO) if end - start == L else None)
    local, readout = torch.cat(local), torch.cat(readout)
    output = []
    for i in range(T):
        c = i // L
        w = chunk_weights(q[i], U[:c], WQ, theta)
        output.append(reverse_history(
            local[i], readout[i], A[:c], B[:c], w,
            skip_zeros=theta is not None,
        ))

    # Preserve the real active prefix, not a padded complete chunk.
    M, n = divmod(T, L)
    cache = dict(A_hist=A[:M], B_hist=B[:M], U_hist=U[:M], n=n)
    if n:
        cache["A"], cache["B"] = A[M], B[M]
    else:
        cache["A"] = torch.eye(Dk, device=k.device, dtype=k.dtype)
        cache["B"] = v.new_zeros(Dk, Dv)
    start = M * L
    full_windows = (n // p) * p
    cache["U"] = list(pool_keys(k[start:start + full_windows], p, WK, WV, WO))
    cache["keys"] = k[start + full_windows:T].clone()
    return torch.stack(output), cache
\end{lstlisting}

\subsection{Dynamic Decode Reference}

Each call receives features for one new token, recomputes its historical
weights, and updates the active cache. The native convolution cache is
maintained separately and is not reset when a chunk is completed.

\begin{lstlisting}[style=pseudocode,breaklines=true]
def hla_decode(q, k, qbar, kbar, v, beta, g, cache, L, p, W, theta):
    # The enclosing layer has already updated its native convolution cache.
    # Inputs describe the new token; theta is an inference-only threshold.
    WQ, WK, WV, WO = W
    Dk = k.numel()
    cache["A"], cache["B"] = affine_step(
        cache["A"], cache["B"], kbar, v, beta, g,
    )
    cache["n"] += 1
    cache["keys"] = torch.cat((cache["keys"], k[None]), dim=0)
    if len(cache["keys"]) == p:
        u = pool_keys(cache["keys"], p, WK, WV, WO)[0]
        cache["U"].append(u)
        cache["keys"] = k.new_empty(0, Dk)

    # Recompute this token's weights; current-chunk weight is one.
    w = chunk_weights(q, cache["U_hist"], WQ, theta)
    r = qbar / math.sqrt(Dk)
    out = r @ cache["B"]
    r = r @ cache["A"]
    out = reverse_history(
        out, r, cache["A_hist"], cache["B_hist"], w,
        skip_zeros=theta is not None,
    )

    # Close the chunk only after computing the current token's output.
    if cache["n"] == L:
        cache["A_hist"].append(cache["A"])
        cache["B_hist"].append(cache["B"])
        cache["U_hist"].append(torch.stack(cache["U"]))
        cache["A"] = torch.eye(Dk, device=k.device, dtype=k.dtype)
        cache["B"] = v.new_zeros(Dk, v.numel())
        cache["U"], cache["n"] = [], 0
    return out, cache
\end{lstlisting}

The listings specify the attention-core forward computation. Router-only
optimization follows Section~\ref{sec:training-objective}; optimizer
steps and auxiliary-loss collection are omitted here. In particular,
inference thresholding is disabled during training.

\subsection{Complexity Analysis}
\label{app:complexity-derivation}

\begin{table}[t]
    \centering
    \caption{
    Per-layer, per-head inference complexity.
    $T$ is the sequence length, $L$ the chunk size,
    $N=\lceil T/L\rceil$ the number of chunks,
    $p$ the pooling-window size, $P=L/p$ the number of
    representatives per chunk, and $M$ the number of accessed chunks
    including the current chunk. We assume $d_k=d_v=d_r=d$.
    }
    \label{tab:complexity-comparison}
    \scriptsize
    \resizebox{\columnwidth}{!}{%
    \begin{tabular}{lcccc}
        \toprule
        Method
        & Prefill
        & Decode / token
        & Inference Cache
        & Chunk Weighting \\
        \midrule
        Softmax Attention
        & $O(T^2d)$
        & $O(Td)$
        & $O(Td)$
        & Query-dependent \\
        Linear Attention
        & $O(Td^2)$
        & $O(d^2)$
        & $O(d^2)$
        & Implicit \\
        GDN
        & $O(Td^2)$
        & $O(d^2)$
        & $O(d^2)$
        & Implicit \\
        MHLA
        & $O(Td^2+N^2d^2)$
        & $O(d^2+Nd^2/L)$
        & $O(Nd^2)$
        & Fixed \\
        \midrule
        \methodname{} (ours)
        & $O(Tpd+TNPd+TMd^2)$
        & $O(pd+NPd+Md^2)$
        & $O(N(d^2+Pd)+Ld)$
        & Query-dependent \\
        \bottomrule
    \end{tabular}}
\end{table}

We analyze inference complexity for one attention layer and head.
Let $T$ be the sequence length, $L$ the chunk size,
$N=\lceil T/L\rceil$, $p$ the pooling-window size, and
$P=L/p$, assuming $d_k=d_v=d_r=d$.
For query $i$, let
$\mathcal{J}_i=\{j<c(i):w_{i,j}\geq\theta\}$
denote the selected historical chunks and
$M_i=1+|\mathcal{J}_i|$ include the current chunk.
For prefill, $M$ denotes the average of $M_i$ over queries.
Common backbone projections are omitted.

\textbf{Baselines.}~
Softmax attention requires $O(T^2d)$ prefill computation,
$O(Td)$ work per decoded token, and $O(Td)$ KV cache.
Recurrent linear attention and GDN maintain a fixed $d\times d$ state,
giving $O(Td^2)$ prefill, $O(d^2)$ decoding, and $O(d^2)$ cache.
For MHLA~\citep{zhang2026mhla}, constructing local states costs
$O(Td^2)$, while combining cached chunk states across query chunks
adds $O(N^2d^2)$ prefill computation. Since the chunk mixture is
shared within each query chunk, its $O(Nd^2)$ construction cost
amortizes to $O(Nd^2/L)$ per token, with $O(Nd^2)$ cache.

\textbf{\methodname{} prefill.}~
HLA constructs exact affine chunk summaries, pooled representatives,
chunk scores, and query-dependent historical readouts.
Affine summaries and local prefix computation require $O(Td^2)$.
Self-attention pooling costs $O(Tpd)$ in total, while scoring at most
$NP$ representatives per query costs $O(TNPd)$.
After thresholding, only $M_i$ transitions are accessed for query $i$,
each requiring $O(d^2)$ vector--matrix computation. Thus,
\begin{equation}
\text{Prefill}
=
O(Tpd+TNPd+TMd^2),
\label{eq:appendix-prefill-complexity}
\end{equation}
where the $O(Td^2)$ term is absorbed by $O(TMd^2)$ since $M\geq1$.

\textbf{\methodname{} decoding.}~
Updating the active affine summary costs $O(d^2)$.
Pooling amortizes to $O(d^2+pd)$ per token, scoring cached
representatives costs $O(NPd)$, and applying the selected transitions
costs $O(Md^2)$. Therefore,
\begin{equation}
\text{Decode}
=
O(pd+NPd+Md^2).
\label{eq:appendix-decode-complexity}
\end{equation}

\textbf{\methodname{} cache.}~
Each completed chunk stores two $d\times d$ affine matrices and
$P$ routing representatives, requiring $O(d^2+Pd)$ storage.
Including the active-chunk buffer gives
\begin{equation}
O\!\left(N(d^2+Pd)+Ld\right).
\label{eq:appendix-cache-complexity}
\end{equation}
All chunk summaries remain cached because a chunk skipped by one query
may be selected by a later query.


\subsection{Analysis of Long-Range Historical Contributions}
\label{app:longrange-analysis}

We provide additional analyses of the from-scratch 1.3B GDN and HLA
models evaluated in Table~\ref{tab:ruler-extrapolation}. Unless otherwise
specified, the contribution analysis uses RULER S1 examples and measures
the RMS norm of the additive contribution from each historical chunk to
the pre-normalization GDN core output. Contributions are normalized over
historical chunks within each layer. This measure characterizes the
relative influence of historical affine transitions and should not be
interpreted as softmax attention or token-level causal attribution.

For the aggregate contribution analyses, we use an unfiltered set of
64 paired examples, consisting of the first 16 generated examples at
each of 4K, 8K, 16K, and 32K. The decomposition is numerically verified
against the original forward computation for every analyzed layer.

\begin{figure*}[t]
    \centering
    \includegraphics[width=0.65\textwidth]
    {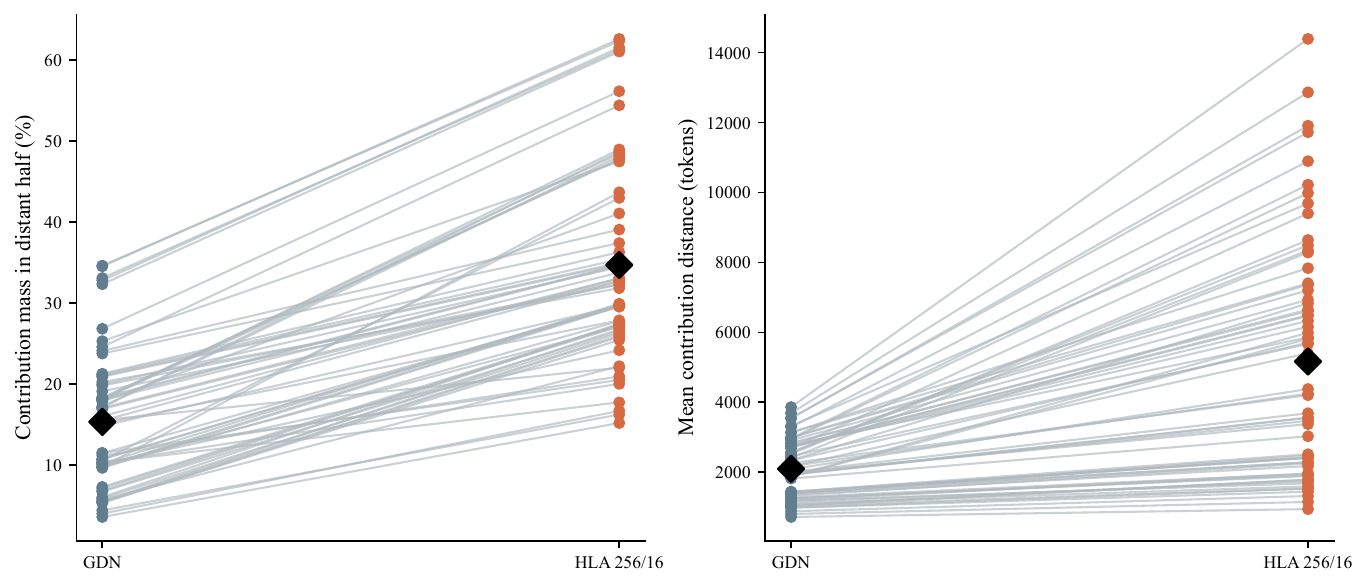}
    \caption{
    \textbf{Aggregate long-range historical contributions.}
    Each line connects the same input under native GDN and HLA.
    Left: fraction of normalized historical contribution assigned to
    chunks in the distant half of the context. Right: contribution-weighted
    mean source distance. Black diamonds denote averages over the
    unfiltered analysis set. HLA shifts historical contribution toward
    more distant context under both measures.
    }
    \label{fig:cohort-contribution-distance}
\end{figure*}

Figure~\ref{fig:cohort-contribution-distance} shows a systematic shift
toward more distant historical information under HLA. Averaged over the
analysis set, the contribution assigned to the distant half of the context
increases from 24.64\% to 42.48\% at 4K, from 16.87\% to 33.09\% at
8K, from 13.58\% to 36.75\% at 16K, and from 6.22\% to 26.44\% at
32K. The corresponding contribution-weighted mean distance also increases
substantially, with the largest gap observed at 32K
(2,876 vs.\ 9,591 tokens).

\begin{figure*}[t]
    \centering
    \includegraphics[width=\textwidth]
    {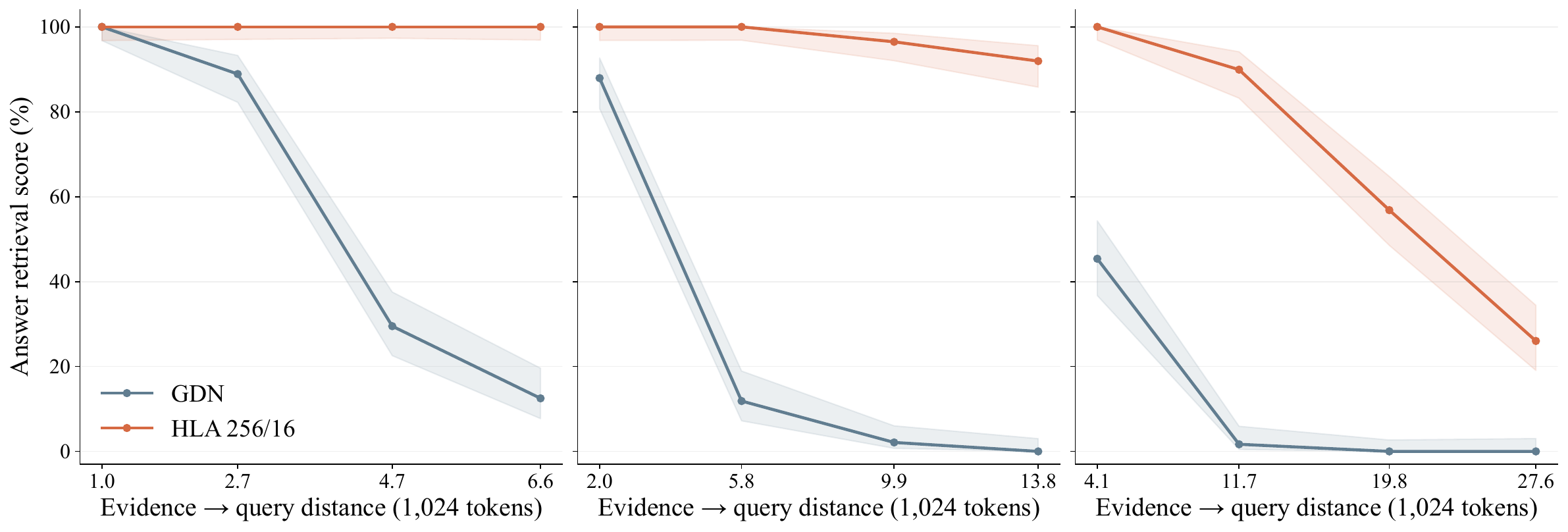}
    \caption{
    \textbf{RULER S1 retrieval performance as a function of evidence distance.}
    Results are grouped by the distance between the target evidence and
    the final query. Curves report answer-retrieval scores, with shaded
    intervals indicating uncertainty within each distance bin.
    HLA degrades substantially more slowly as the evidence moves farther
    from the query, particularly at 16K and 32K.
    }
    \label{fig:retrieval-by-distance}
\end{figure*}

\begin{figure*}[t]
    \centering
    \includegraphics[width=\textwidth]
    {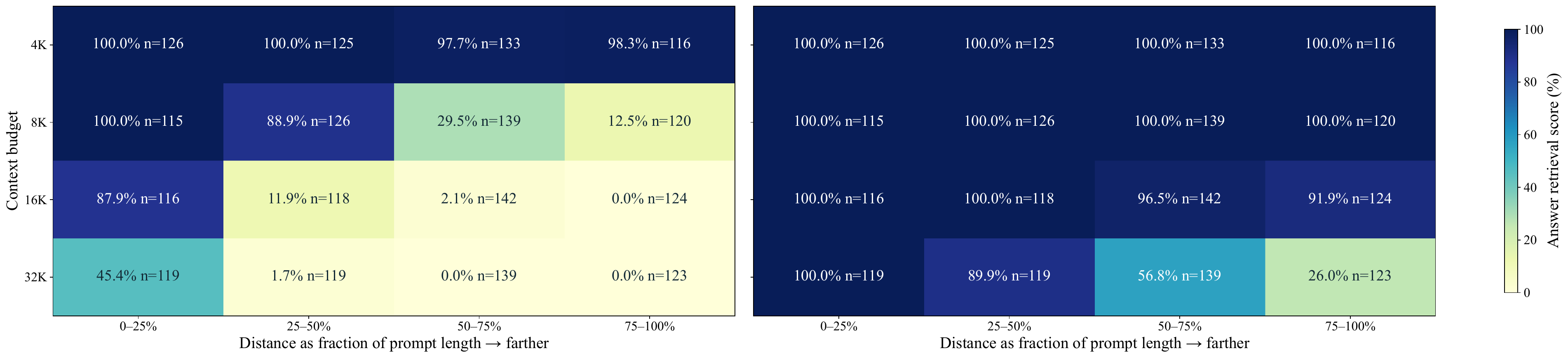}
    \caption{
    \textbf{Retrieval performance across context lengths and relative evidence positions.}
    Each cell reports the RULER S1 retrieval score for examples grouped
    by the evidence-to-query distance as a fraction of the prompt length.
    Left: native GDN. Right: HLA. HLA retains substantially higher
    retrieval performance for distant evidence as the context length
    increases.
    }
    \label{fig:retrieval-distance-heatmap}
\end{figure*}

Figures~\ref{fig:retrieval-by-distance} and
\ref{fig:retrieval-distance-heatmap} directly relate retrieval quality
to evidence distance. Native GDN deteriorates rapidly as the target
moves farther from the query, whereas HLA preserves substantially higher
retrieval accuracy. At 16K, for example, HLA remains above 90\% even in
the farthest distance bin, while GDN approaches zero. At 32K, the same
trend persists, although both models become more challenging at the
largest distances. These results are consistent with the stronger
long-range historical contributions observed above.

\begin{figure*}[p]
    \centering
    \includegraphics[width=0.85\textwidth]
    {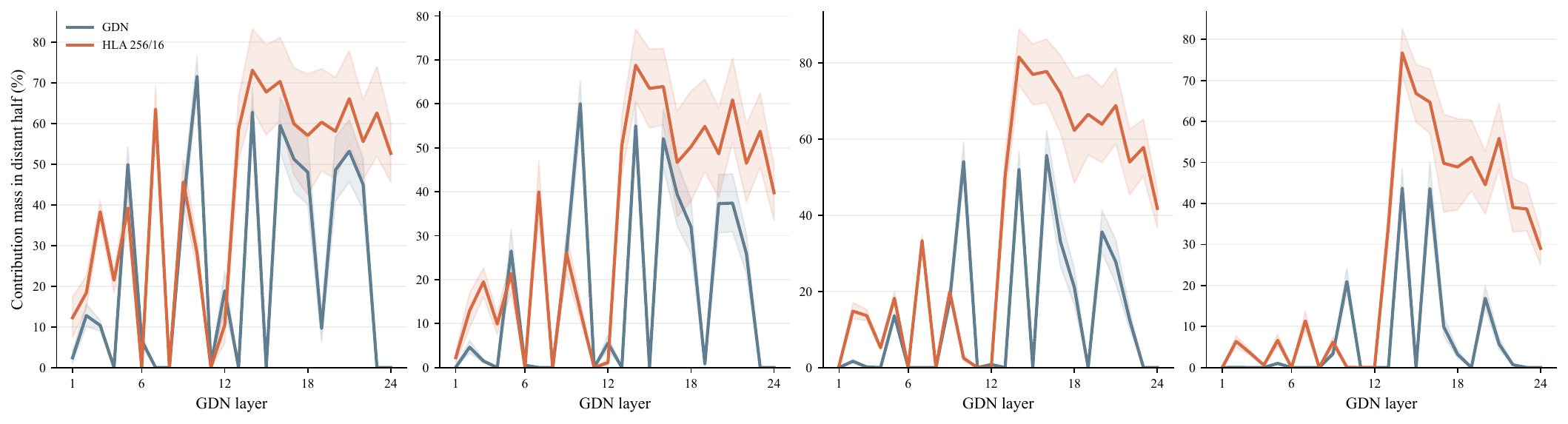}
    \caption{
    \textbf{Layer-wise contribution from distant history.}
    From left to right, the panels correspond to 4K, 8K, 16K, and 32K.
    Each curve reports the fraction of normalized historical contribution
    assigned to the distant half of the context, averaged over paired
    examples. Shaded regions denote pointwise 95\% bootstrap intervals.
    HLA exhibits substantially stronger distant-history contributions in
    several middle and upper layers, with the difference becoming more
    pronounced at longer contexts.
    }
    \label{fig:layer-far-mass}
\end{figure*}

\begin{figure*}[p]
    \centering
    \includegraphics[width=0.85\textwidth]
    {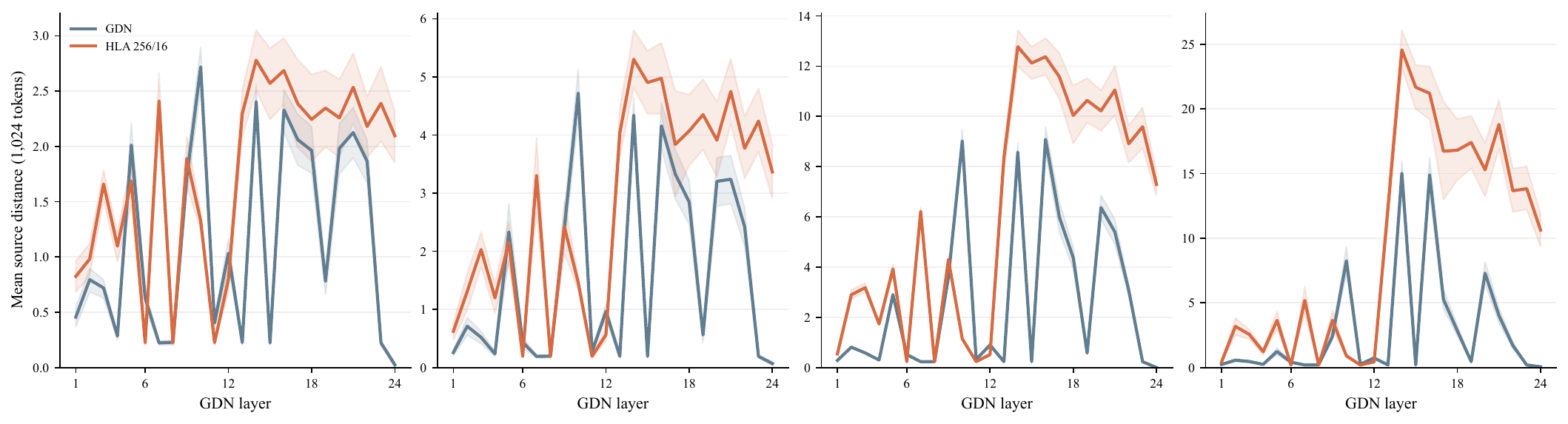}
    \caption{
    \textbf{Layer-wise contribution-weighted source distance.}
    From left to right, the panels correspond to 4K, 8K, 16K, and 32K.
    The metric measures the average distance of historical sources,
    weighted by their normalized effective contributions. HLA increases
    the contribution distance in several middle and upper layers,
    particularly at 16K and 32K.
    }
    \label{fig:layer-mean-distance}
\end{figure*}

\begin{figure*}[p]
    \centering
    \includegraphics[width=0.85\textwidth]
    {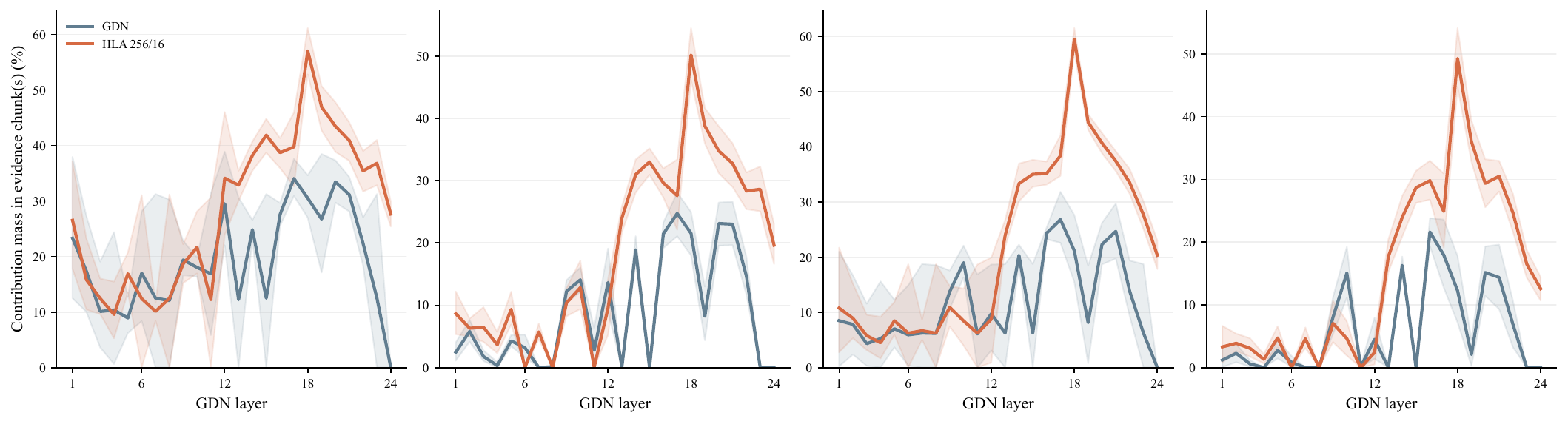}
    \caption{
    \textbf{Layer-wise contribution assigned to answer-evidence chunks.}
    From left to right, the panels correspond to 4K, 8K, 16K, and 32K.
    We report the normalized historical contribution assigned to the
    chunk(s) containing the target evidence. HLA assigns greater relative
    contribution to evidence-containing chunks in many middle and upper
    layers, especially at longer context lengths.
    }
    \label{fig:layer-evidence-mass}
\end{figure*}

Figures~\ref{fig:layer-far-mass}--\ref{fig:layer-evidence-mass} show that
the difference between HLA and native GDN is strongly layer-dependent.
The largest changes occur primarily in a subset of middle and upper
layers rather than uniformly throughout the network. In these layers,
HLA both extends the effective source distance and assigns greater
relative contribution to distant and evidence-containing chunks. This
suggests that query-dependent routing induces structured changes in how
different layers access historical information rather than simply
amplifying all distant context.

\subsubsection{Qualitative Contribution Maps}
\label{app:qualitative-contribution-maps}

\begin{figure*}[p]
    \centering
    \includegraphics[width=0.94\textwidth]
    {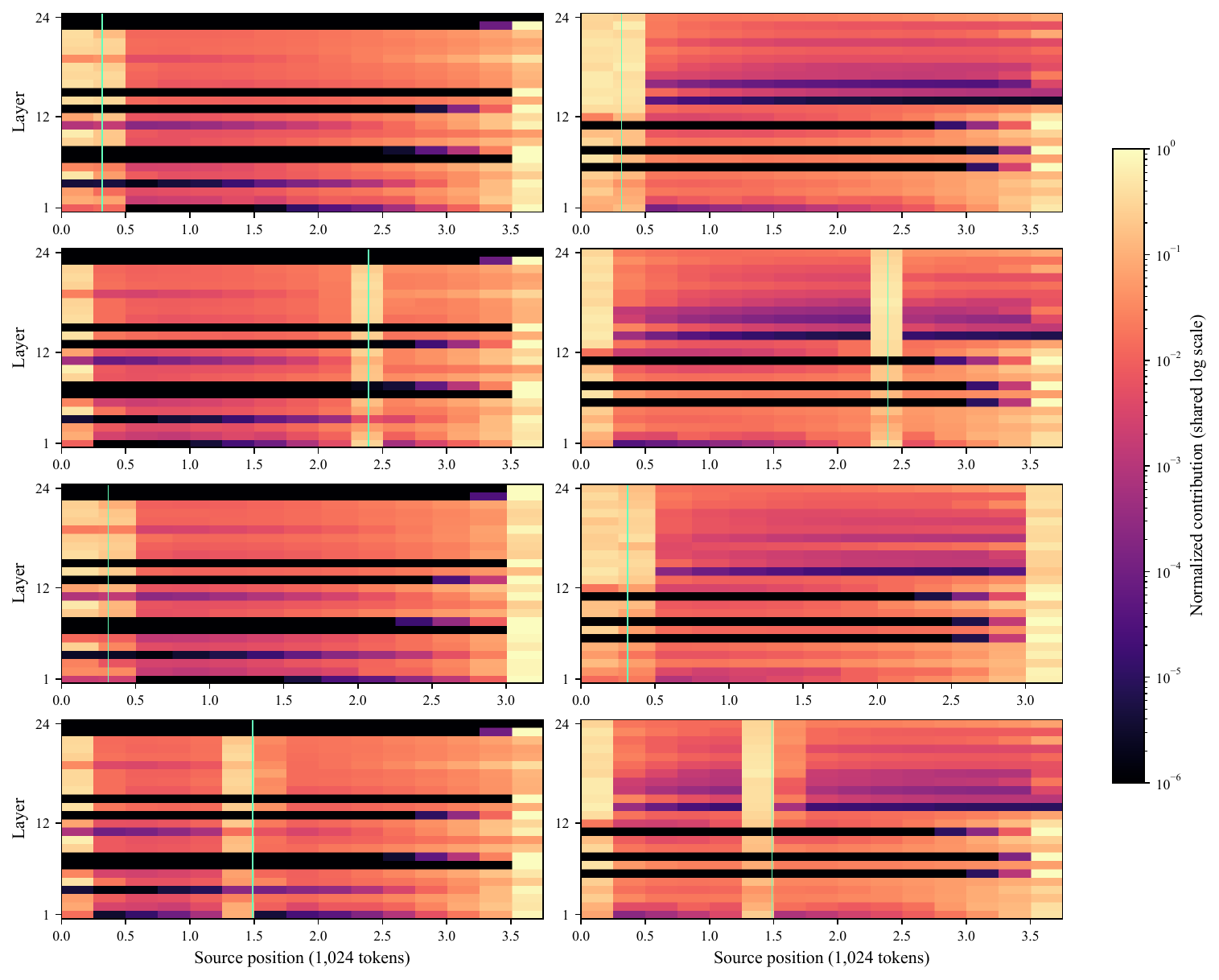}

    \vspace{0.8em}

    \includegraphics[width=0.94\textwidth]
    {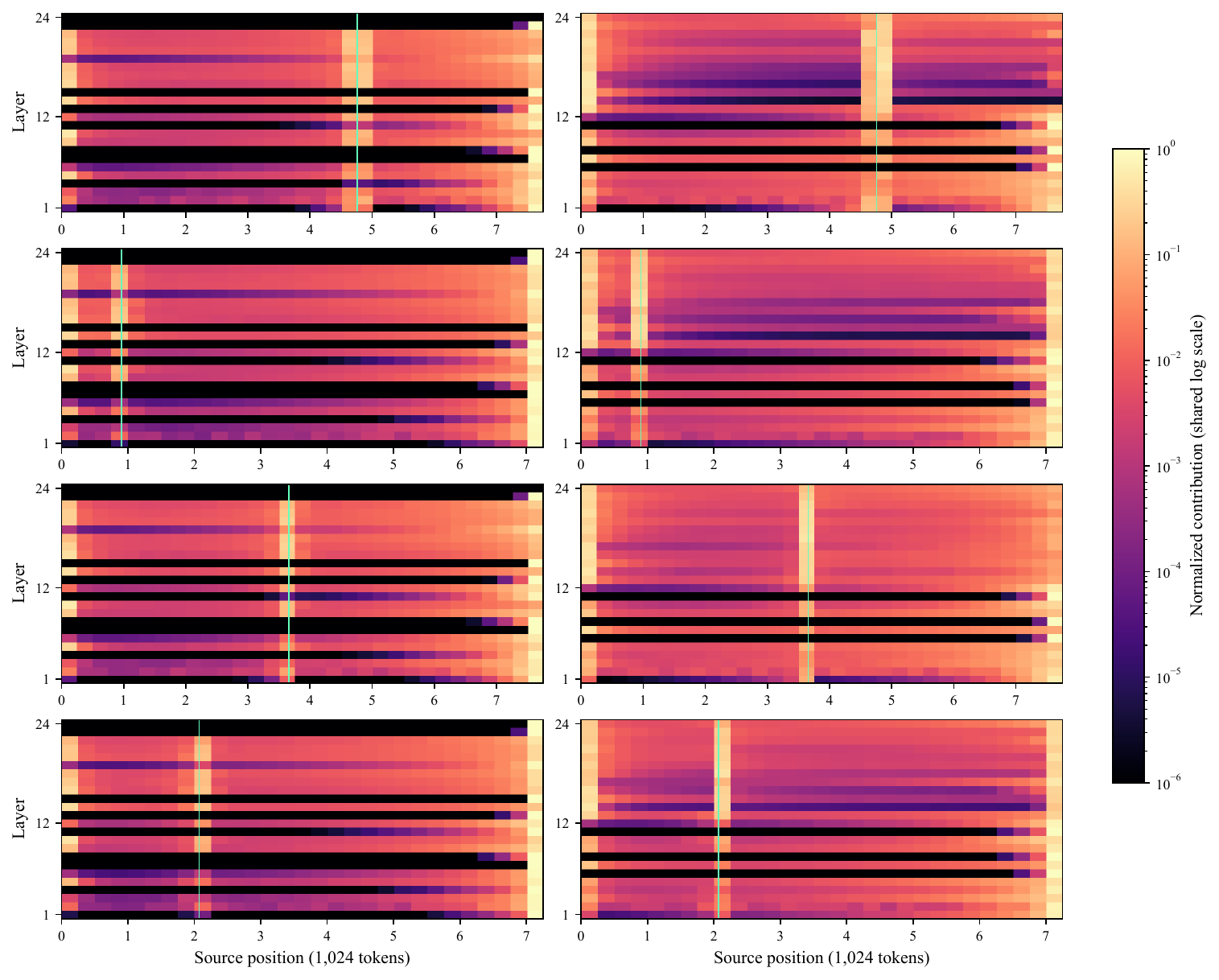}

    \caption{
    \textbf{Historical-contribution maps at 4K and 8K context lengths.}
    Each panel compares native GDN (left) and HLA (right) on the same
    RULER S1 examples. Rows correspond to different inputs and the
    vertical marker denotes the target-evidence position. HLA generally
    distributes non-negligible contribution over a broader range of
    historical positions.
    }
    \label{fig:contribution-overview-short}
\end{figure*}

\begin{figure*}[p]
    \centering
    \includegraphics[width=0.94\textwidth]
    {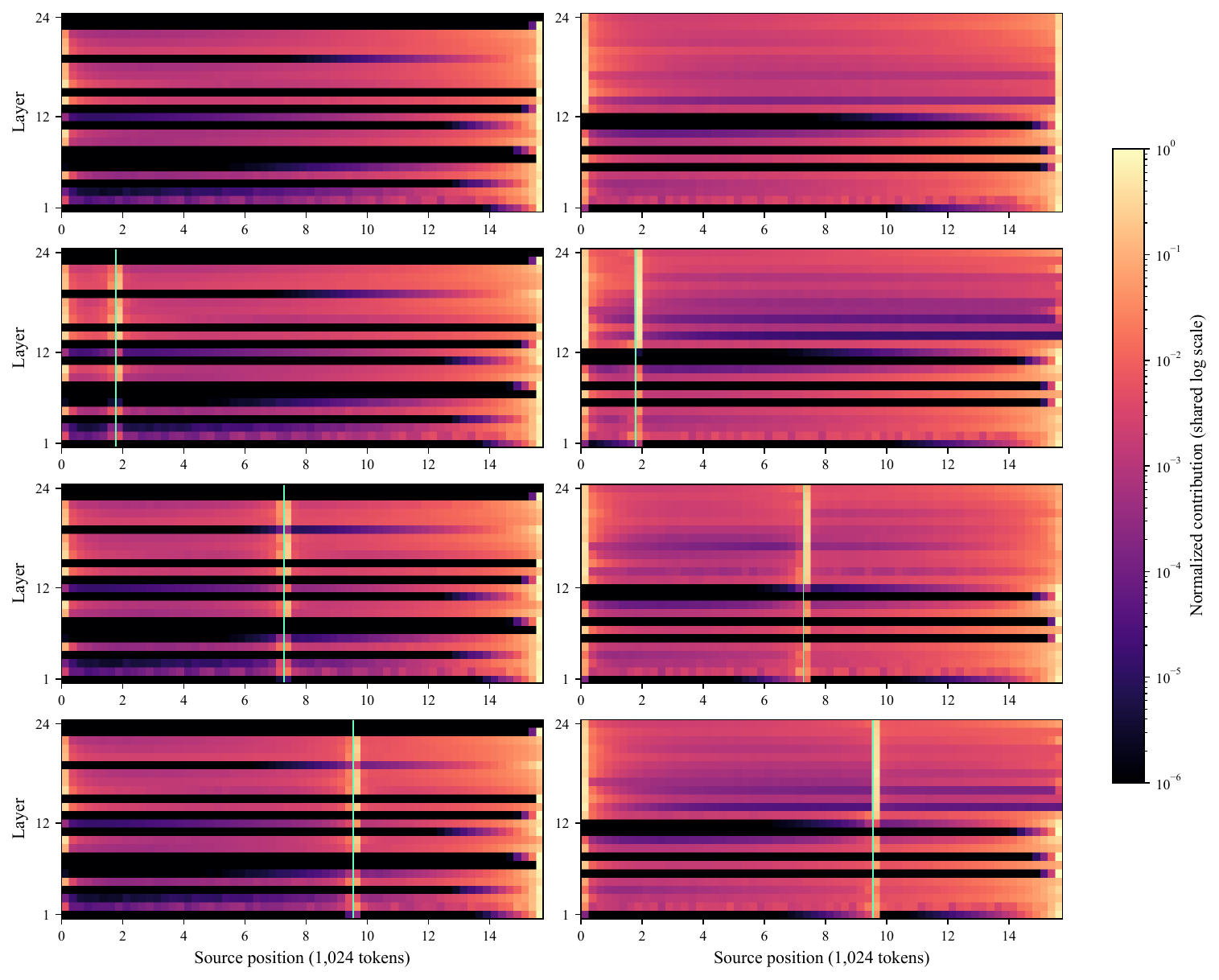}

    \vspace{0.8em}

    \includegraphics[width=0.94\textwidth]
    {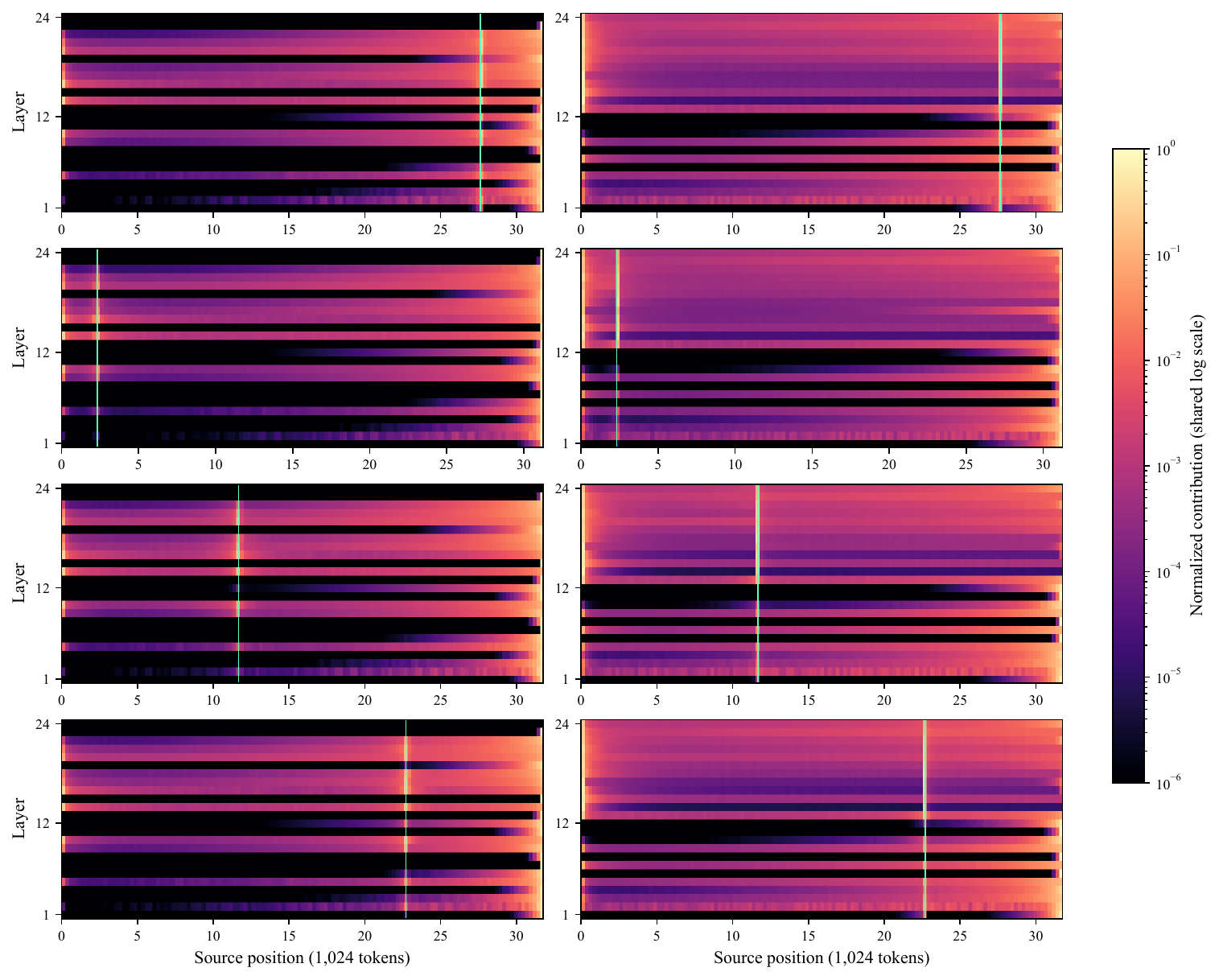}

    \caption{
    \textbf{Historical-contribution maps at 16K and 32K context lengths.}
    The visualization follows Figure~\ref{fig:contribution-overview-short}.
    At longer contexts, native GDN increasingly concentrates its
    historical contribution near the most recent positions, whereas HLA
    retains visible contributions from substantially more distant history,
    including regions around the target evidence in several examples.
    These visualizations are qualitative and do not constitute token-level
    causal attribution.
    }
    \label{fig:contribution-overview-long}
\end{figure*}

We further visualize normalized historical contributions for individual
RULER S1 examples as shown in Figures~\ref{fig:contribution-overview-short}--\ref{fig:contribution-overview-long}. Each row corresponds to a paired input, with native
GDN on the left and HLA on the right. The horizontal axis denotes source
position and the vertical axis denotes model depth. The vertical marker
indicates the location of the target evidence. A shared logarithmic scale
is used to expose weak but non-zero historical contributions.

\end{document}